%% file: arxiv.tex
\RequirePackage{fix-cm}
\documentclass{article}
\usepackage{PRIMEarxiv}
\usepackage[utf8]{inputenc}
\usepackage[T1]{fontenc}
\usepackage[english]{babel}
\usepackage{microtype}

\input{math_commands.tex}

\usepackage{float}
\usepackage[authoryear,round]{natbib}
\usepackage{hyperref}
\usepackage{url}

\usepackage{graphicx}
\usepackage{wrapfig}
\usepackage{tikz}
\usetikzlibrary{arrows.meta,calc,fit,positioning,patterns}
\usepackage{pgfplots}
\pgfplotsset{compat=1.18}
\usepackage{xcolor}
\usepackage{listings}
\usepackage{needspace}
\usepackage{enumitem}

\definecolor{figInk}{HTML}{27323A}
\definecolor{figMuted}{HTML}{69757E}
\definecolor{figRule}{HTML}{D5DBDF}
\definecolor{figPaper}{HTML}{F7F8F7}
\definecolor{figTeal}{HTML}{277C74}
\definecolor{figTealLight}{HTML}{E7F2F0}
\definecolor{figCoral}{HTML}{C45F4D}
\definecolor{figCoralLight}{HTML}{F8ECE8}
\definecolor{figAmber}{HTML}{A97822}
\definecolor{figAmberLight}{HTML}{FAF2DF}
\definecolor{figBlue}{HTML}{486F95}
\definecolor{figBlueLight}{HTML}{E9F0F6}

\lstdefinestyle{benchmarkprompt}{
    basicstyle=\ttfamily\footnotesize,
    columns=fullflexible,
    keepspaces=true,
    breaklines=true,
    breakatwhitespace=true,
    breakindent=0pt,
    showstringspaces=false,
    frame=single,
    rulecolor=\color{figRule},
    backgroundcolor=\color{figPaper},
    framesep=5pt,
    xleftmargin=5pt,
    xrightmargin=5pt,
    aboveskip=0.6em,
    belowskip=0.8em
}

\pgfplotsset{
    bench axis/.style={
        axis x line*=bottom,
        axis y line*=left,
        y axis line style={draw=none},
        ytick style={draw=none},
        tick align=outside,
        tick style={draw=figMuted, line width=0.45pt},
        xmajorgrids,
        grid style={draw=figRule, line width=0.4pt},
        axis line style={draw=figMuted, line width=0.55pt},
        tick label style={font=\scriptsize, text=figInk},
        label style={font=\small, text=figInk},
        title style={font=\small\bfseries, text=figInk},
        legend style={font=\scriptsize, draw=none, fill=none, text=figInk},
        clip=false
    }
}

\hypersetup{
    colorlinks=true,
    linkcolor=figTeal,
    citecolor=figTeal,
    urlcolor=figBlue,
    pdftitle={TCSAlgBench: Benchmarking Automated Proving for Research-Level Theoretical Computer Science},
    pdfauthor={Chutong Yang, Xiyuan Zhang, Yu Huang, Boran Han, Soonho Kong, Shuai Zhang, Vihang Prakash Patil, Zhen Han, Michael Bohlke-Schneider, Bernie Wang}
}

\title{TCSAlgBench: Benchmarking Automated Proving for Research-Level Theoretical Computer Science}
\author{%
  \begin{tabular}{c}
    Chutong Yang$^{1,}$\thanks{Work done during internship at Amazon.},
    Xiyuan Zhang$^{2,}$%
    \thanks{Correspondence to: Xiyuan Zhang (\texttt{xiyuanz@amazon.com}).}, Yu Huang$^{3,*}$,
    Boran Han$^{2}$, Soonho Kong$^{2}$ \\[2pt]
    Shuai Zhang$^{2}$, Vihang Prakash Patil$^{2}$, Zhen Han$^{2}$,
    Michael Bohlke-Schneider$^{2}$, Bernie Wang$^{2}$ \\[6pt]
    {\small $^{1}$Department of Computer Science, The University of Texas at Austin} \\
    {\small $^{2}$Amazon \qquad $^{3}$Department of Statistics and Data Science, The Wharton School, University of Pennsylvania}
  \end{tabular}%
}
\date{}

\makeatletter
\renewcommand{\@maketitle}{%
  \vbox{%
    \hsize\textwidth
    \linewidth\hsize
    \vskip 0.1in
    \@toptitlebar
    \centering
    {\LARGE\scshape\@title\par}%
    \@bottomtitlebar
    \vskip 0.12in
    {% Give adjacent author-note markers their natural width.
      \renewcommand{\@makefnmark}{\hbox{\@textsuperscript{\normalfont\@thefnmark}}}%
      \normalsize\@author\par
    }%
    \vskip 0.12in
  }%
}
\makeatother

\begin{document}
\maketitle

\begin{abstract}
Large language models perform strongly on competition mathematics, but their research-level reasoning remains difficult to evaluate systematically. Theoretical computer science (TCS) connects algorithm design to explicit guarantees and fundamental limits, providing a setting for evaluating whether models can justify computational improvements with arguments humans can inspect. We introduce TCSAlgBench, a benchmark and reusable pipeline for natural-language proof discovery, comprising 398 theorem-level challenges from 138 STOC and COLT 2026 papers. Expert-designed rules complete paper-specific context, preserve computational assumptions and quantitative guarantees, and withhold constructions when discovering an algorithm is part of the task. For each task, prover systems receive theorem statements and access to cited prior work. The pipeline supports fresh, versioned challenge batches from newly released papers. We evaluate ten model configurations from four families under direct inference and prover-verifier discussion, and compare four agent workflows under matched model-call opportunities. All evaluations use the full benchmark. In the model comparison, GPT-5.6 Sol max achieves the highest five-run verifier-accepted coverage at 23.6\% after 10-round discussion. Discussion and repeated sampling improve coverage. In the separate agent comparison using GPT-5.5 xhigh, decomposition improves coverage over discussion, and agentic planning achieves the highest five-run verifier-accepted coverage at 25.4\%. TCSAlgBench provides a refreshable testbed for measuring progress in model reasoning and studying how agent workflows support research-level proof discovery.
\end{abstract}

% The substantive sections are shared with the conference version.
\begingroup
\setlength{\abovecaptionskip}{6pt}
\setlength{\belowcaptionskip}{0pt}
% Give captions a little more separation from surrounding body text.
\setlength{\textfloatsep}{14pt}
\setlength{\intextsep}{14pt}
\setlength{\floatsep}{12pt}
\raggedbottom
\input{sections/introduction}
\input{sections/related_work}
\input{sections/tasks_and_research_uses}
\input{sections/illustrative_evaluation}
\input{sections/conclusion}
\endgroup

\subsection*{AI use statement}
In this work, we used generative AI tools to generate synthetic data sets, implement methods, design or provide feedback on research methodology or experiments, clean and reformat the dataset.
We have not used generative AI tools to propose or refine hypotheses, support qualitative and thematic data analysis, or interpret results,
and help develop theoretical models or conceptual frameworks, formulate mathematical claims, provide critical ingredients for proving mathematical claims, assist in the writing of proofs, assist with translation, are not applicable to this work.
Additionally, we used generative AI tools to refine the paper writing and literature research. We have reviewed all AI-assisted work. We have reviewed all the writing and citations and tested the AI-generated code. We are responsible for the research idea and use AI only to refine our experimental design. We take responsibility for the final content of this work, including text, claims, or artifacts produced with the aid of generative AI.

\subsection*{Reproducibility statement}
We will release 166 challenges on Hugging Face at \url{https://huggingface.co/datasets/cyang98/TCSAlgBENCH}. These challenges are derived from 57 papers whose source versions are licensed under CC BY 4.0 or CC0. Further, we will release the titles of all 138 papers we used in our paper, with the TeX files publicly available on arXiv. With the pipeline prompt in Appendix~\ref{sec:prompt-templates}, it should be easy to reproduce the challenges used in our paper. The prover workflow we used in our paper is also easy to reproduce using the prover prompts in Appendix~\ref{sec:prompt-templates}.

\bibliography{iclr2027_conference}
\bibliographystyle{iclr2027_conference}

\appendix
% math_commands.tex uses \reg for regularization; this example uses it for regret.
\begingroup
\renewcommand{\reg}{\mathrm{Reg}}
\let\arxivsubsection\subsection
\renewcommand{\subsection}{\Needspace{8\baselineskip}\arxivsubsection}
\input{sections/appendix_theorem-upper-bound}
\endgroup
\input{sections/appendix_dataset_construction_and_validation}
\input{sections/appendix_evaluation_protocol}
\input{sections/appendix_secondary_results}

\input{sections/appendix_prompts}
\end{document}

%% file: math_commands.tex
\usepackage{amsmath,amsfonts,bm}

\def\eqref#1{equation~\ref{#1}}
\def\1{\bm{1}}

\DeclareMathAlphabet{\mathsfit}{\encodingdefault}{\sfdefault}{m}{sl}
\SetMathAlphabet{\mathsfit}{bold}{\encodingdefault}{\sfdefault}{bx}{n}

\newcommand{\E}{\mathbb{E}}

\newcommand{\reg}{\lambda}

%% file: sections/introduction.tex
\section{Introduction}

% Limit wrapping to page 1 so page 2 resumes at full text width.
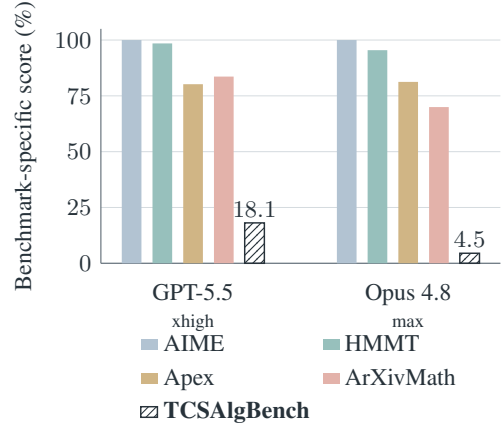
\begin{wrapfigure}[19]{r}{0.4\textwidth}
    % Align the wrapped figure so its caption stays inside the first-page text area.
    \vspace{\dimexpr-\intextsep-\baselineskip\relax}
    \centering
    \input{figures/benchmark_comparison}\unskip\par\nointerlineskip
    %\caption{MathArena scores and TCSAlgBench five-run coverage with 10-round discussion: a proof accepted in at least one of five runs ($n=398$). See Appendix~\ref{app:benchmark-comparison}.}
    \caption{Strong published final-answer accuracy on selected MathArena benchmarks contrasts with low proof coverage on TCSAlgBench, even after five runs of 10-round discussion. }
    \label{fig:benchmark-comparison}
\end{wrapfigure}

Large language models (LLMs) have achieved significant improvements in mathematics, from the International Mathematical Olympiad (IMO) to research-level problems. However, many existing benchmarks focus on competitive mathematics, while recent research-level achievements often focus on individual problems, making it difficult to systematically evaluate research-level capabilities or compare agent workflows. A broader goal for LLM research is to develop models that help improve computational systems while giving humans a basis for understanding and evaluating those improvements. Theoretical computer science (TCS) connects algorithms to explicit assumptions, mathematical guarantees, and fundamental limits. In areas such as privacy and fairness, this reasoning makes precise which property a proposed method satisfies, when the claim holds, and what tradeoffs it entails. Progress in TCS reasoning could therefore support LLMs as research collaborators whose proposals come with justifications that humans can inspect and challenge. Evaluating proof discovery in TCS tests a capability needed for this vision: reasoning about algorithms, the information they can access, and their data and computational requirements.

We introduce \emph{TCSAlgBench}, a benchmark and construction pipeline for generating self-contained natural-language TCS statements to evaluate LLM reasoning ability, with a focus on algorithm design and sample- or runtime-complexity analysis. TCSAlgBench has three properties that make it useful for evaluating research-level reasoning. (1) \textbf{Diversity.} Its initial collection contains 398 theorem-level challenges from 138 STOC and COLT 2026 papers, spanning learning theory, privacy, optimization, sampling, and other areas of TCS. Each challenge provides a target theorem together with its local definitions, assumptions, notation, and necessary prior work, and asks for a rigorous natural-language proof. (2) \textbf{Generalization.} The reusable pipeline can generate fresh, versioned challenge batches from newly released papers, allowing successive models and agent designs to be evaluated on problems published after their assumed knowledge cutoffs. The size and topic composition of each batch can be controlled, reducing reliance on a fixed benchmark that may be contaminated over time. (3) \textbf{Difficulty.} Current models achieve low proof coverage on TCSAlgBench even across five runs of 10-round discussion, despite strong published final-answer accuracy on selected math benchmarks~\citep{dekoninck2026matharena}; Figure~\ref{fig:benchmark-comparison} illustrates this gap.

%Expert-designed rules preserve computational models, interaction protocols, and quantitative guarantees while removing constructions or intermediate results that reveal the intended solution. When discovering an algorithm is part of the challenge, its construction is withheld. Completing missing definitions helps prevent solvers from adopting interpretations that inadvertently simplify the task. We developed the extraction and repair rules on 21 challenges and manually reviewed 61 final statements, including the development set and 40 additional challenges, for fidelity and self-containment. All headline natural-language model and workflow comparisons use the full collection of 398 challenges. We provide the prompts used in the pipeline in Appendix~\ref{sec:prompt-templates}.

A key contribution of our pipeline is a set of \emph{expert-designed rules} for turning extracted theorems into fair proof-discovery problems. These rules capture details that generic theorem extraction can miss: they preserve computational models, restrictions on information access and action order, assumptions, and quantitative guarantees, while adding the definitions needed to interpret the statement. They also remove source-paper lemmas and, when algorithm design is part of the task, rewrite named
algorithms as existence claims so that the construction is not revealed. Applied in a fixed sequence to the paper's proof-dependency graph, the rules produce self-contained problems without leaking the central argument. We developed the rules on 21 challenges and manually reviewed 61 final statements, including the development set and 40 additional challenges, for fidelity and self-containment. All headline natural-language model and workflow comparisons use the full collection of 398 challenges. We provide the prompts used in the pipeline in Appendix~\ref{sec:prompt-templates}.

% The evaluation uses an offline sandbox with a fixed library of cited prior work, focusing on proof discovery after the research question and relevant literature have been supplied. We compare model configurations under direct inference, repeated sampling, and prover-verifier refinement, and evaluate four proof-agent workflows using verifier acceptance (Section~\ref{sec:main-evaluation-protocol}). The workflow comparison matches model-call opportunities and reports realized input and output tokens separately. Agentic planning reaches the highest five-run coverage, while MCTS reaches the highest seed-1 acceptance. We assess verifier sensitivity by re-scoring identical submissions with Opus~4.8.

Our evaluation measures both sources of progress in automated proof generation: the capability of the underlying model and the design of the agent built around it. Every model configuration and agent workflow is evaluated on all 398 challenges in an offline sandbox with the same fixed library of cited prior work. For model capability, we ask: Which model should TCS researchers use? In this work, we compare ten configurations from four model families under direct inference and multi-round prover--verifier discussion, with repeated runs in both settings. For agent design, we ask: Which workflow design should TCS researchers use? We compare four representative workflows that cover the main mechanisms used by proof agents: iterative refinement, explicit decomposition, tree search, and adaptive planning (Section~\ref{sec:main-evaluation-protocol}). The agent comparison holds the backbone model, problem input, tool access, model-call opportunities, and per-call generation limit fixed, while reporting realized input and output tokens separately. We report both seed-1 acceptance and five-run coverage: MCTS performs best on the former, while agentic planning performs best on the latter. Finally, we re-score identical submissions with Opus~4.8 to assess how sensitive these conclusions are to the choice of verifier. As a supplementary resource, we provide 221 provisional Lean statements that pass compilation and automated semantic screening for future formalization work.

Our main contributions are:
\begin{itemize}[nosep,leftmargin=4mm]
    \item \textbf{A scalable, refreshable benchmark-construction framework.} We introduce an end-to-end pipeline that transforms recent research papers into self-contained proof challenges while preserving computational models,
    restrictions on information access and action order, and complexity guarantees. The pipeline supports versioned, post-cutoff evaluation on newly released work.
    \item \textbf{A broad research-level TCS benchmark.} We construct 398 theorem-level challenges from 138 STOC and COLT 2026 papers spanning major areas of TCS. Each task isolates genuine proof discovery by providing only the necessary local context and cited prior work.
    \item \textbf{A controlled study of models and proof-agent design.} We evaluate ten model configurations and four representative agent workflows across the full benchmark, controlling call opportunities and reporting repeated-run coverage, token costs,
    and verifier sensitivity. The results expose substantial capability gaps and quantify the trade-offs among discussion, decomposition, tree search, and agentic planning.
\end{itemize}

%% file: figures/benchmark_comparison.tex
% Published MathArena model-page snapshot: 2026-09-18.
% Exact values, sources, and protocol differences: app:benchmark-comparison.
% Each model group contains the same four answer benchmarks and our proof task.
% TCSAlgBench: five-run coverage of 10-round discussion, 72/398 and 18/398.
\begingroup
\colorlet{benchAIME}{figBlue!42!white}
\colorlet{benchHMMT}{figTeal!48!white}
\colorlet{benchApex}{figAmber!55!white}
\colorlet{benchArxiv}{figCoral!48!white}
\begin{tikzpicture}
    \begin{axis}[
        name=benchmarkBars,
        scale only axis,
        width=0.80\linewidth,
        height=3.1cm,
        xmin=0, xmax=13,
        ymin=0, ymax=105,
        xtick={3,10},
        xticklabels={{GPT-5.5\\{\scriptsize xhigh}},{Opus 4.8\\{\scriptsize max}}},
        xticklabel style={align=center},
        ytick={0,25,50,75,100},
        ylabel={Benchmark-specific score (\%)},
        label style={font=\footnotesize, text=figInk},
        tick label style={font=\footnotesize, text=figInk},
        axis lines*=left,
        axis line style={draw=figMuted, line width=0.45pt},
        tick style={draw=none},
        ymajorgrids,
        grid style={draw=figRule, line width=0.35pt},
        ybar,
        bar width=7.4pt,
        bar shift=0pt,
        clip=false,
        legend columns=2,
        legend cell align=left,
        legend image code/.code={\path[#1] (0cm,-0.07cm) rectangle (0.22cm,0.07cm);},
        legend style={at={(0.5,-0.28)}, anchor=north,
            font=\footnotesize, text=figInk, draw=none, fill=none,
            /tikz/every even column/.append style={column sep=3pt},
            /tikz/row sep=1pt, inner sep=0pt},
    ]
        \addplot[fill=benchAIME, draw=none] coordinates {(1,100) (8,100)};
        \addlegendentry{AIME}
        \addplot[fill=benchHMMT, draw=none] coordinates {(2,98.48) (9,95.45)};
        \addlegendentry{HMMT}
        \addplot[fill=benchApex, draw=none] coordinates {(3,80.21) (10,81.25)};
        \addlegendentry{Apex}
        \addplot[fill=benchArxiv, draw=none] coordinates {(4,83.63) (11,69.97)};
        \addlegendentry{ArXivMath}
        \addplot[pattern=north east lines, pattern color=figInk,
            draw=figInk, line width=0.55pt,
            nodes near coords={\pgfmathprintnumber[fixed,precision=1,zerofill]{\pgfplotspointmeta}},
            every node near coord/.append style={font=\footnotesize\bfseries,
                text=figInk, anchor=south, yshift=2pt, inner sep=0pt},
        ] coordinates {(5,18.1) (12,4.5)};
        \addlegendentry{\textbf{TCSAlgBench}}
    \end{axis}
\end{tikzpicture}
\endgroup

%% file: sections/related_work.tex
\vspace{-0.5em}
\section{Related Work}\label{sec:related_work}
\vspace{-0.5em}
\paragraph{Mathematical and TCS benchmarks.}
Mathematical evaluations span exact-answer datasets~\citep{cobbe2021gsm8k,hendrycks2021math}, difficult or fresh problems and false-premise tests~\citep{phan2025humanitys,balunovic2025matharena,petrovDV25}, and research-level problems~\citep{glazerEtAl24frontiermath,schmittEtAl25improofbench,abouzaid2026first,abouzaidSWW26}. Advances from competition mathematics and discovery~\citep{hubertEtAl26alphaproof,novikov2025alphaevolve,fengEtAl26,anthropic2026zeta,alon2026unitdistance} to even the Navier–Stokes problem~\citep{openai2026navierstokes}, alongside reported formalizations~\citep{anthropic2026fermat}, motivate systematic evaluation. LemmaBench extracts self-contained lemmas from arXiv papers and supports recurring updates~\citep{peyronnetGH26}. Concurrent TCS-BENCH provides 300 natural-language tasks from STOC, FOCS, and SODA, with dependency-based masking and expert-evaluated judging~\citep{cohenAddadEtAl26tcsbench}. FormalTCS offers expert-validated instances with natural-language and Lean statements and proofs~\citep{wangEtAl26formaltcs}. TCSAlgBench emphasizes theorem-level TCS proof discovery, TCS-specific context repairs, and an offline library of cited prior work. It withholds source-paper lemmas and, for algorithm-design tasks, named constructions. Differences are in task inputs and pipeline design.

\paragraph{Proof-agent designs.}
Proof agents combine iterative generation and critique~\citep{fengEtAl26,anYPZ26,schmittEtAl26proofcouncil, requenaLNBS26}, sketch-based decomposition~\citep{jiangEtAl23dsp,zhangEtAl25cumulative,varamballyEtAl25hilbert, renEtAl25deepseekv2}, AND--OR or value-guided search~\citep{lampleEtAl22htps,xinEtAl25deepseek,tsoukalasEtAl26alphaproofnexus}, and replanning over dependency structures~\citep{anYPZ26,wuEtAl26starpolya,chungEtAl26goedelarchitect}. General agent benchmarks emphasize versioned tasks, external evaluation, and resource accounting~\citep{merrillEtAl26,sunEtAl26,liEtAl26}. Theorem-dependency graphs support claim generation and retrieval~\citep{busbibW26,kurganWLSARAII26}; DeFAb benchmarks defeasible abduction~\citep{cooperV26}. Section~\ref{sec:experimental-setup} explains how workflows implemented in this work represent these mechanisms.

\paragraph{LLM-based proof verification.}
LLM judging has demonstrated substantial agreement with human judgments~\citep{zheng2023judging}. For mathematical proofs, the Open Proof Corpus reports strong agreement with expert labels~\citep{dekoninckEtAl25opc}, while ProofGrader improves scoring through reference solutions, rubrics, and ensembling~\citep{maEtAl26proofgrader}. Pseudo-Formalization rewrites proofs into independently checked premise--conclusion modules, improving error-localization tradeoffs on competition and research proofs~\citep{barkallahEtAl26pseudoformal}. TCS-BENCH supplies its judge with a reference proof~\citep{cohenAddadEtAl26tcsbench}; our pipeline checks cited statements against their sources before judging the submitted proof.

\paragraph{Autoformalization and formal proving.}
Autoformalization spans early translation studies and paired corpora~\citep{wuEtAl22autoformalization,azerbayev2023proofnet, ying2024leanworkbook,patelEtAl23arxiv2formal}, TCS collections~\citep{zhangEtAl25leanmeetstcs,fengEtAl26lcsbench}, and textbook and research formalization~\citep{rammalPGHKMAC26,moakhar2026beyond,zhangEtAl26leanmarathon,liuZZDHLWXZWZLL26}. LeanDojo supports formal proof agents~\citep{yang2023leandojo}; miniF2F and PutnamBench provide machine-checked evaluation~\citep{zheng2022minif2f,tsoukalas2024putnambench}. Compilation does not ensure statement fidelity; model consensus is an imperfect, human-calibrated proxy~\citep{zhang2026beyondcompilation}.

%% file: sections/tasks_and_research_uses.tex
\section{Benchmark Scope and Evaluation}\label{sec:benchmark}

TCSAlgBench evaluates natural-language proof discovery for TCS claims involving algorithm design and sample- or runtime-complexity guarantees. It focuses on the stage of research after the problem and relevant literature have been identified by human researchers or other agents. We try to mimic the real research process, where human researchers propose a problem and provide previous work using their domain knowledge or other agents. 

\subsection{Benchmark Construction and Pipeline Development}\label{sec:benchmark-construction}

The evaluation collection contains 398 challenges from 138 publicly available arXiv papers accepted to STOC and COLT 2026. Every challenge records its source and version. The public release includes challenges derived from arXiv versions licensed under CC BY 4.0 or CC0. A model-assisted screen retains papers whose main results give upper or lower bounds on sample or runtime complexity. The collection spans algorithmic fairness and calibration, differential privacy, learning theory, optimization, sampling, and other TCS areas, such as cryptography, graph theory, and coding theory. These areas underpin fair decision-making and calibrated prediction, privacy-preserving data analysis, learning from limited data, scalable optimization, and sampling for probabilistic inference. Cryptography, graph algorithms, and coding theory further support secure communication, routing and matching, and reliable data transmission. Our focus on sample and runtime complexity results tests a prerequisite for LLMs to contribute to such applications: reasoning about which computational improvements are achievable and at what cost. Upper-bound challenges can require constructing algorithms and proving their correctness and efficiency, while lower-bound challenges require establishing unavoidable limitations under explicit assumptions. These tasks examine whether models can support proposed methods with checkable arguments and identify constraints that human researchers must consider when assessing their usefulness.

\begin{figure}[t]
    \centering
    \resizebox{\textwidth}{!}{%
        \input{figures/challenge_pipeline_overview}%
    }
    \par\nointerlineskip
    \caption{Constructing proof-discovery challenges. The source proof graph is used only for selection. LLM checks guide source-text insertion to complete the context. Repair examples are schematic; existence rewrites preserve assumptions and quantitative guarantees when algorithm design is part of the task. Appendix~\ref{sec:challenge-construction} details the construction passes.}
    \label{fig:challenge-pipeline}
\end{figure}
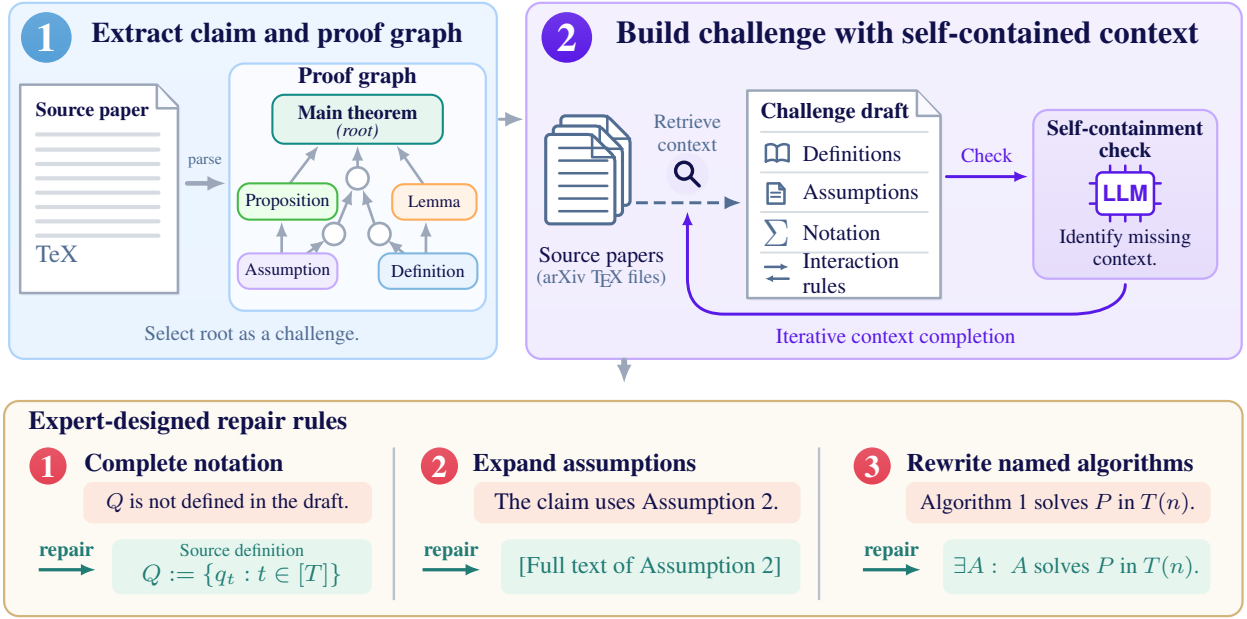

Figure~\ref{fig:challenge-pipeline} summarizes the construction pipeline, which uses Opus~4.8 with high reasoning mode. We resolve each source paper and the cited prior work needed to interpret it, parse the TeX sources into proof graphs of statements, definitions, and dependencies, and select one to four main theorems per source paper. Code assembles the selected source statements into a draft, and an LLM self-containment checker guides context completion.

%Expert-designed repair rules are applied automatically to resolve internal references and interaction protocols, harmonize notation, and remove intermediate results or construction details that reveal the intended solution. 
A central methodological contribution is our expert-designed repair layer, which bridges the gap between theorem extraction and fair, self-contained proof discovery. Applied automatically in a fixed sequence, its rules resolve internal references, recover rules governing information access and action order, harmonize notation, and remove intermediate results or construction details that expose the intended solution.
For example, when algorithm design is part of the task, a claim about ``Algorithm~1'' is rewritten as an existence claim that preserves its assumptions and quantitative guarantees. Mathematical context is quoted from the source, subject to notation normalization and these documented rewrites. Added standard-background definitions are explicitly labeled.

We developed the extraction, context completion, and repair rules using 21 challenges across topics. After the final repair pass, we manually reviewed 61 challenge statements (these 21 development challenges and 40 additional challenges across topics) for fidelity and self-containment. All 40 additional challenges were judged faithful to the source results and self-contained. All headline model and workflow comparisons use the full collection of \mbox{398 challenges}: the 61 reviewed statements, including the development set, and the remaining 337 challenges. Appendix~\ref{sec:construction} details construction, statement review, and source-date diagnostics.

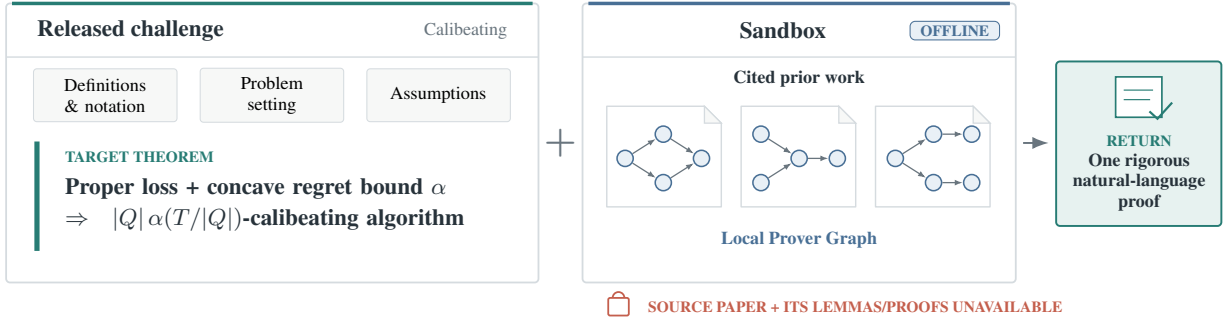
\begin{figure}[t]
    \centering
    \resizebox{0.98\textwidth}{!}{%
        \input{figures/challenge_anatomy}%
    }
    \par\nointerlineskip
    \caption{Proof-discovery evaluation interface. A challenge supplies the target theorem with its definitions, notation, assumptions, and problem setting. The offline sandbox permits access to cited prior work through the local prover graphs and the TeX file. The system returns one natural-language proof. Appendix~\ref{app:calibeating-challenge} shows a complete challenge.}
    \label{fig:challenge-anatomy}
\end{figure}

\subsection{Evaluation Protocol}\label{sec:main-evaluation-protocol}

A prover system receives a self-contained theorem statement with its definitions, assumptions, notation, and problem setting, and returns one proof in ordinary mathematical prose. The evaluator-controlled offline sandbox provides a fixed corpus of cited prior work through the local prover graphs, graph representations of those papers and their results. The prover may use results from this corpus. The source paper, its construction graph, and artifacts containing the target proof are excluded. Source-paper lemmas are neither supplied as additional facts nor inlined into the challenge; the prover must establish any such claims needed for its argument. The public release represents the cited documents through a versioned manifest of source URLs, versions, and checksums. Figure~\ref{fig:challenge-anatomy} illustrates this interface using the Calibeating challenge reproduced in full in Appendix~\ref{app:calibeating-challenge}.

Our primary outcome is \emph{verifier acceptance}. Before scoring, a proof-organizing step checks whether statements cited from prior work match their original source statements. Three separately sampled GPT-5.5 high-effort voters receive only the problem statement and rewritten proof, with acceptance requiring at least two PASS votes. Internal verifier feedback guides search but does not determine the benchmark label. To assess sensitivity to the choice of judge, we re-score identical submissions with Opus~4.8 high effort. An author additionally reviewed 10 accepted proofs produced by GPT-5.5 xhigh through prover-verifier discussion on challenges in the manually reviewed statement set. All 10 accepted proofs were correct.

We report \emph{single-run acceptance} (seed~1) and \emph{multi-run coverage}, which counts a challenge once if any independently seeded run is accepted. The workflow comparison fixes the backbone model, problem input, and tool access, and matches model-call opportunities through shared limits on iterations, proof-goal attempts, and discussion rounds. The per-call generation limit is also fixed. Input and output tokens are reported separately because workflows can process different amounts of context within the same call budget. Section~\ref{sec:experimental-setup} describes the workflows, and Appendix~\ref{sec:evaluation_protocol} defines call accounting, outcome aggregation, and confidence scoring.

\paragraph{LLM-based verification.}
LLMs are commonly used for proof critique and automated assessment in recent mathematical reasoning systems and benchmarks, with evaluations against expert judgments supporting this practice~\citep{dekoninckEtAl25opc,maEtAl26proofgrader}. Our proof-agent and verification prompts adapt BrokenMath~\citep{petrovDV25} and QED~\citep{anYPZ26}, incorporating checks for false premises, unsupported citations, and unresolved proof obligations. Our rubric also shares criteria with concurrent TCS-BENCH~\citep{cohenAddadEtAl26tcsbench}, including logical rigor, faithful use of assumptions and cited results, and quantitative correctness. Automated judging enables repeated comparisons on fresh, versioned batches, with a fixed judging protocol within each batch and targeted expert audits. Appendix~\ref{sec:verification-precedents} summarizes the empirical evidence and protocol differences; Appendix~\ref{sec:prompt-verifier} provides our prompts.

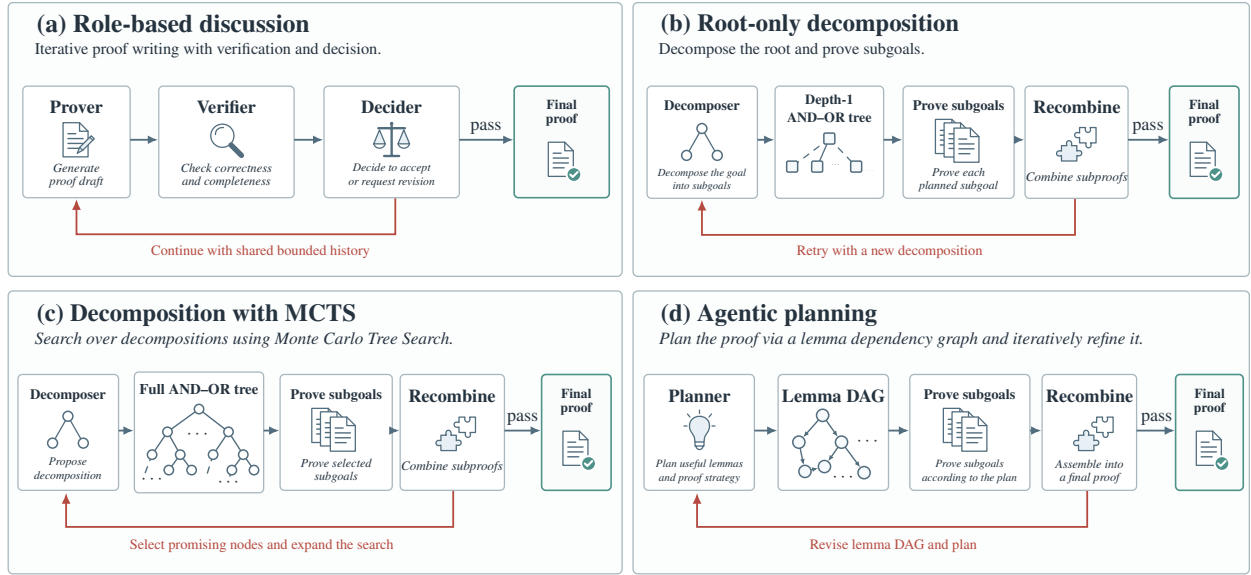
\begin{figure}[t]
    \centering
    \resizebox{\textwidth}{!}{%
        \input{figures/agent_workflows}%
    }
    \par\nointerlineskip
    \caption{Information flow in the four call-matched agent designs. Rightward arrows lead to a final proof on an internal stop or pass; red return arrows indicate retries. Final submissions are scored separately by the external verifier. Discussion reuses a shared history; root-only decomposition retries depth-1 reductions; MCTS searches a full AND--OR tree; and agentic planning revises a lemma DAG from critic feedback.}
    \label{fig:agent-workflows}
\end{figure}

\subsection{Provisional Lean Statements}\label{sec:lean-formalization}

Formalizing selected TCS results is feasible, but broader coverage requires specialized definitions and supporting libraries~\citep{wangEtAl26formaltcs,rammalPGHKMAC26}. As a supplementary result, we contribute a reusable statement-formalization pipeline for the community, producing candidate Lean~4 propositions in a pinned Mathlib/CSLib environment. Candidates must compile without \texttt{sorry}, \texttt{admit}, or untrusted axioms. Two checks require approval from both GPT-5.5 and Opus~4.8 high-effort reviewers: \emph{consistency} with a back-translation generated without the challenge, and \emph{degeneracy} checks for vacuous or trivial encodings. We retain 221 candidate statements (55.5\% of 398 challenges) using our pipeline. Appendix~\ref{sec:construction} details construction and topic counts; Appendix~\ref{app:secondary} reports the results.

%% file: figures/challenge_pipeline_overview.tex
% Editable vector figure following the user-supplied reference style.
% Stage 2 shows context retrieval and the checker feedback loop.
\begingroup%
\definecolor{refNavy}{HTML}{10104B}%
\definecolor{refBlue}{HTML}{559FD7}%
\definecolor{refBlueLine}{HTML}{AED4FC}%
\definecolor{refBlueFill}{HTML}{EAF4FF}%
\definecolor{refPurple}{HTML}{5919E6}%
\definecolor{refPurpleLine}{HTML}{C5AAFF}%
\definecolor{refPurpleFill}{HTML}{F1EDFE}%
\definecolor{refMuted}{HTML}{526D92}%
\definecolor{refRule}{HTML}{9CA8B9}%
\definecolor{refPaperLine}{HTML}{D7DCE3}%
\definecolor{refTeal}{HTML}{138A86}%
\definecolor{refRepair}{HTML}{217C75}%
\definecolor{refWarm}{HTML}{FDFAF3}%
\definecolor{refWarmLine}{HTML}{D5B47C}%
\definecolor{refBefore}{HTML}{FCE8DF}%
\definecolor{refAfter}{HTML}{E6F6EE}%
\definecolor{refRed}{HTML}{E44355}%
\begin{tikzpicture}[
    x=0.01cm, y=-0.01cm, text=refNavy,
    font=\fontsize{8.5}{9.4}\selectfont,
    panel/.style={rounded corners=5pt, line width=0.75pt},
    textnode/.style={inner sep=0pt, align=center},
    document/.style={draw=refRule, fill=white,
        line width=0.85pt, line join=round},
    paperline/.style={draw=refPaperLine, line width=1.4pt, line cap=round},
    flow/.style={-{Latex[length=2.2mm,width=1.8mm]},
        draw=refRule, line width=1.55pt, line cap=round},
    dependency/.style={-{Latex[length=1.8mm,width=1.4mm]},
        draw=refRule, line width=0.9pt},
    feedback/.style={-{Latex[length=1.65mm,width=1.3mm]},
        draw=refPurple, line width=1.0pt, line cap=round},
    repairflow/.style={-{Latex[length=1.8mm,width=1.4mm]},
        draw=refRepair, line width=1.1pt},
    glyph/.style={draw=refMuted, line width=0.85pt,
        line cap=round, line join=round},
    contextlabel/.style={anchor=west, inner sep=0pt,
        font=\fontsize{7.5}{8.3}\selectfont},
    connectorlabel/.style={textnode, font=\fontsize{6.8}{7.6}\selectfont},
    repairheading/.style={anchor=west, inner sep=0pt,
        font=\fontsize{8.5}{9.2}\selectfont\bfseries}
]
    \path[use as bounding box] (80,80) rectangle (1560,823);

    % Reference panels and numbered stage headings.
    \draw[panel, draw=refBlueLine, top color=refBlueFill!85!white,
        bottom color=refBlueFill] (90,88) rectangle (668,510);
    \draw[panel, draw=refPurpleLine, top color=refPurpleFill!80!white,
        bottom color=refPurpleFill] (703,88) rectangle (1551,510);
    \shade[top color=refBlue!90!white,bottom color=refBlue]
        (134,129) circle (30);
    \node[textnode, text=white, font=\fontsize{13.5}{14}\selectfont\bfseries]
        at (134,129) {1};
    \node[anchor=west, inner sep=0pt] at (187,126)
        {\resizebox{4.40cm}{!}{\bfseries Extract claim and proof graph}};
    \shade[top color=refPurple!87!white,bottom color=refPurple]
        (751,129) circle (30);
    \node[textnode, text=white, font=\fontsize{13.5}{14}\selectfont\bfseries]
        at (751,129) {2};
    \node[anchor=west, inner sep=0pt] at (809,126)
        {\resizebox{6.90cm}{!}{\bfseries Build challenge with self-contained context}};

    % Source paper, including its folded corner and source description.
    \path[fill=refNavy, opacity=0.045]
        (106,188) -- (268,188) -- (293,213) -- (293,435) -- (106,435) -- cycle;
    \draw[document] (103,185) -- (265,185) -- (290,210)
        -- (290,432) -- (103,432) -- cycle;
    \draw[document, fill=refBlueFill!35!white]
        (265,185) -- (265,213) -- (290,213);
    \node[anchor=west, inner sep=0pt] at (121,216)
        {\resizebox{1.34cm}{!}{\bfseries Source paper}};
    \foreach \lineY in {244,262,279,296,313,330,347,364} {
        \draw[paperline] (123,\lineY) -- (267,\lineY);
    }
    \node[anchor=north west, inner sep=0pt, text=refMuted]
        at (121,376)
        {\resizebox{0.50cm}{!}{TeX}};
    \draw[flow] (301,302) -- (347,302);
    \node[textnode, text=refMuted] at (322,277)
        {\resizebox{0.40cm}{!}{parse}};

    % Detailed proof graph from the reference.
    \draw[panel, draw=refBlueLine, fill=white, fill opacity=0.70]
        (351,160) rectangle (655,453);
    \node[textnode] at (503,176)
        {\resizebox{1.41cm}{!}{\bfseries Proof graph}};
    \draw[rounded corners=4pt, draw=refTeal, fill=refAfter, line width=0.7pt]
        (402,197) rectangle (604,255);
    \node[textnode] at (503,218)
        {\resizebox{1.42cm}{!}{\bfseries Main theorem}};
    \node[textnode] at (503,240)
        {\resizebox{0.50cm}{!}{\itshape(root)}};

    % DAG edges point from supporting statements toward dependent claims.
    % Every edge moves upward, giving a strict topological ordering.
    \draw[dependency] (422,303) -- (457,257);
    \draw[dependency] (581,303) -- (549,257);
    \draw[dependency] (503,283) -- (503,257);
    \draw[dependency] (413,385) -- (413,347);
    \draw[dependency] (584,385) -- (584,347);
    \draw[dependency] (480,350) -- (497,309);
    \draw[dependency] (523,350) -- (509,309);
    \draw[dependency] (443,385) -- (465,369);
    \draw[dependency] (558,385) -- (538,369);
    \foreach \nodeX/\nodeY in {503/296,475/362,529/362} {
        \draw[draw=refRule, fill=white, line width=0.9pt]
            (\nodeX,\nodeY) circle (13);
    }
    \draw[rounded corners=4pt, draw=green!45!gray, fill=green!8,
        line width=0.7pt] (361,302) rectangle (479,345);
    \node[textnode] at (420,323.5)
        {\resizebox{1.00cm}{!}{Proposition}};
    \draw[rounded corners=4pt, draw=orange!65, fill=orange!10,
        line width=0.7pt] (544,303) rectangle (644,345);
    \node[textnode] at (594,324)
        {\resizebox{0.62cm}{!}{Lemma}};
    \draw[rounded corners=4pt, draw=refPurpleLine, fill=refPurpleFill,
        line width=0.7pt] (361,385) rectangle (479,427);
    \node[textnode] at (420,406)
        {\resizebox{1.01cm}{!}{Assumption}};
    \draw[rounded corners=4pt, draw=refBlue!75, fill=refBlueFill,
        line width=0.7pt] (528,385) rectangle (645,427);
    \node[textnode] at (586.5,406)
        {\resizebox{0.87cm}{!}{Definition}};
    \node[textnode, text=refMuted] at (378,482)
        {\resizebox{2.55cm}{!}{Select root as a challenge.}};

    % Panel-to-panel transition from Stage 1 to Stage 2.
    \draw[flow, line width=1.0pt] (671,226) -- (700,226);

    \begin{scope}[shift={(24,0)}]
    % Three stacked folded source documents.
    \foreach \paperX/\paperY in {720/222,711/234,702/244} {
        \draw[draw=refMuted, fill=white, rounded corners=1.5pt,
            line width=1.05pt]
            (\paperX,\paperY) -- ({\paperX+55},\paperY)
            -- ({\paperX+82},{\paperY+27})
            -- ({\paperX+82},{\paperY+106})
            -- (\paperX,{\paperY+106}) -- cycle;
        \draw[glyph] ({\paperX+55},\paperY)
            -- ({\paperX+55},{\paperY+27})
            -- ({\paperX+82},{\paperY+27});
    }
    \foreach \lineY in {260,275,290,305,320,335} {
        \draw[glyph, line width=0.85pt] (718,\lineY) -- (775,\lineY);
    }
    \node[textnode, font=\fontsize{7.5}{8.3}\selectfont]
        at (768,389) {Source papers};
    \node[textnode, text=refMuted, font=\fontsize{6.5}{7.3}\selectfont]
        at (768,415) {(arXiv \TeX{} files)};

    % Context retrieval cue and magnifying glass.
    \node[connectorlabel, text=refMuted] at (870,242)
        {Retrieve\\context};
    \path[fill=refPurple!10] (870,291) circle (25);
    \draw[draw=refNavy, line width=1.15pt] (866,287) circle (10);
    \draw[draw=refNavy, line width=1.15pt, line cap=round]
        (874,295) -- (884,305);
    \draw[-{Latex[length=2mm,width=1.6mm]}, draw=refMuted,
        dashed, dash pattern=on 4pt off 2pt, line width=1.0pt]
        (809,325) -- (934,325);

    % Folded context sheet and four line icons.
    \path[fill=refNavy, opacity=0.04]
        (944,195) -- (1146,195) -- (1173,224) -- (1173,441)
        -- (944,441) -- cycle;
    \draw[document] (941,192) -- (1142,192) -- (1170,221)
        -- (1170,438) -- (941,438) -- cycle;
    \draw[document, fill=refPurpleFill!35!white]
        (1142,192) -- (1142,222) -- (1170,222);
    \node[anchor=west, inner sep=0pt] at (956,218)
        {\resizebox{1.76cm}{!}{\bfseries Challenge draft}};
    \foreach \lineY in {241,289,336,384} {
        \draw[paperline, line width=0.65pt] (957,\lineY) -- (1155,\lineY);
    }
    \node[contextlabel] at (1006,267) {Definitions};
    \node[contextlabel] at (1006,314) {Assumptions};
    \node[contextlabel] at (1006,360) {Notation};
    \node[contextlabel, align=left] at (1006,408) {Interaction\\rules};
    % Book.
    \draw[glyph] (963,276) -- (963,257)
        .. controls (969,254) and (974,256) .. (976,259)
        .. controls (980,256) and (985,254) .. (990,257)
        -- (990,276)
        .. controls (984,274) and (980,273) .. (976,276)
        .. controls (972,273) and (968,274) .. (963,276);
    \draw[glyph] (976,259) -- (976,276);
    % Assumption sheet.
    \draw[glyph] (965,302) -- (979,302) -- (985,309)
        -- (985,326) -- (965,326) -- cycle;
    \draw[glyph] (979,302) -- (979,309) -- (985,309);
    \draw[glyph, line width=0.65pt]
        (969,313) -- (980,313) (969,318) -- (980,318);
    \node[textnode, text=refMuted, font=\fontsize{13}{13}\selectfont]
        at (975,361) {$\Sigma$};
    % Opposing arrows for interaction rules.
    \draw[glyph, -{Latex[length=0.95mm,width=0.85mm]}]
        (962,401) -- (990,401);
    \draw[glyph, -{Latex[length=0.95mm,width=0.85mm]}]
        (990,415) -- (962,415);

    % Compact checker with a model-neutral LLM chip icon.
    \draw[panel, draw=refPurpleLine, top color=refPurple!10!white,
        bottom color=refPurple!15!white]
        (1280,215) rectangle (1498,415);
    \node[textnode] at (1389,249)
        {\resizebox{1.85cm}{!}{\bfseries\shortstack{Self-containment\\check}}};
    \foreach \pinX in {1371,1389,1407} {
        \draw[glyph, draw=refPurple]
            (\pinX,281) -- (\pinX,289)
            (\pinX,339) -- (\pinX,347);
    }
    \foreach \pinY in {302,326} {
        \draw[glyph, draw=refPurple]
            (1347,\pinY) -- (1355,\pinY)
            (1423,\pinY) -- (1431,\pinY);
    }
    \draw[draw=refPurple, fill=white, rounded corners=2pt, line width=0.85pt]
        (1355,289) rectangle (1423,339);
    \node[textnode, text=refPurple,
        font=\sffamily\bfseries\fontsize{7.5}{8}\selectfont]
        at (1389,314) {LLM};
    \node[textnode, font=\fontsize{6.8}{7.5}\selectfont]
        at (1389,378) {Identify missing\\context.};
    \draw[feedback] (1177,294) -- (1274,294);
    \node[connectorlabel, text=refPurple] at (1225,269) {Check};

    % One continuous return path: missing context guides retrieval and insertion.
    \draw[feedback] (1389,422)
        .. controls (1389,440) and (1368,457) .. (1324,457)
        -- (914,457)
        .. controls (884,457) and (870,442) .. (870,414)
        -- (870,335);
    \node[textnode, text=refPurple, font=\fontsize{7.2}{8}\selectfont]
        at (1117,485) {Iterative context completion};
    \end{scope}

    % Downward connection to the full-width repair panel.
    \draw[flow, line width=2pt] (819,512) -- (819,539);
    \draw[panel, draw=refWarmLine, fill=refWarm]
        (84,560) rectangle (1552,815);
    \node[anchor=west, inner sep=0pt] at (112,586)
        {\resizebox{3.78cm}{!}{\bfseries Expert-designed repair rules}};
    \begin{scope}[shift={(0,-35)}]
    \draw[draw=refRule!70, line width=0.8pt] (546,650) -- (546,831);
    \draw[draw=refRule!70, line width=0.8pt] (1050,650) -- (1050,831);

    % Three red example badges.
    \foreach \badgeX/\badgeNumber in {136/1,600/2,1113/3} {
        \shade[top color=refRed!90!white,bottom color=refRed]
            (\badgeX,672) circle (22);
        \node[textnode, text=white, font=\fontsize{10.5}{11}\selectfont\bfseries]
            at (\badgeX,672) {\badgeNumber};
    }

    % Example 1.
    \node[repairheading] at (178,671) {Complete notation};
    \path[rounded corners=3pt, fill=refBefore]
        (178,691) rectangle (520,740);
    \node[textnode] at (349,716)
        {\resizebox{2.88cm}{!}{$Q$ is not defined in the draft.}};
    \path[rounded corners=3pt, fill=refAfter]
        (212,759) rectangle (520,824);
    \node[textnode, text=refRepair, font=\fontsize{6.1}{6.8}\selectfont]
        at (366,772) {Source definition};
    \node[textnode, text=refRepair] at (366,801)
        {\resizebox{2.37cm}{!}{$Q:=\{q_t:t\in[T]\}$}};
    \node[textnode, text=refRepair] at (158,773)
        {\resizebox{0.65cm}{!}{\bfseries repair}};
    \draw[repairflow] (125,795) -- (193,795);

    % Example 2.
    \node[repairheading] at (639,671) {Expand assumptions};
    \path[rounded corners=3pt, fill=refBefore]
        (641,691) rectangle (1028,740);
    \node[textnode] at (834,716)
        {\resizebox{3.37cm}{!}{The claim uses Assumption 2.}};
    \path[rounded corners=3pt, fill=refAfter]
        (673,759) rectangle (1028,824);
    \node[textnode, text=refRepair] at (850,792)
        {\resizebox{3.12cm}{!}{[Full text of Assumption 2]}};
    \node[textnode, text=refRepair] at (614,773)
        {\resizebox{0.65cm}{!}{\bfseries repair}};
    \draw[repairflow] (580,795) -- (649,795);

    % Example 3.
    \begin{scope}[shift={(12,0)}]
    \node[repairheading] at (1141,671) {Rewrite named algorithms};
    \path[rounded corners=3pt, fill=refBefore]
        (1141,691) rectangle (1502,740);
    \node[textnode] at (1321,716)
        {\resizebox{3.26cm}{!}{Algorithm 1 solves $P$ in $T(n)$.}};
    \path[rounded corners=3pt, fill=refAfter]
        (1185,759) rectangle (1502,824);
    \node[textnode, text=refRepair] at (1343,792)
        {\resizebox{2.92cm}{!}{$\exists A:\ A$ solves $P$ in $T(n)$.}};
    \node[textnode, text=refRepair] at (1124,773)
        {\resizebox{0.65cm}{!}{\bfseries repair}};
    \draw[repairflow] (1090,795) -- (1159,795);
    \end{scope}
    \end{scope}
\end{tikzpicture}%
\endgroup%

%% file: figures/challenge_anatomy.tex
\begin{tikzpicture}[
    x=1cm,
    y=1cm,
    font=\scriptsize,
    flow/.style={-Latex, draw=figMuted, line width=0.75pt},
    dagflow/.style={-{Latex[length=1.1mm,width=0.8mm]}, draw=figMuted, line width=0.5pt},
    dagnode/.style={circle, minimum size=0.22cm, inner sep=0pt,
        draw=figBlue, line width=0.6pt, fill=figBlueLight}
]
    % Self-contained challenge.
    \draw[draw=figRule, line width=0.7pt, fill=white, rounded corners=1.5pt]
        (0,0.65) rectangle (7.25,4.45);
    \draw[figTeal, line width=1.35pt] (0.08,4.45) -- (7.17,4.45);
    \node[anchor=west, font=\small\bfseries, text=figInk] at (0.30,4.08)
        {Released challenge};
    \node[anchor=east, font=\scriptsize, text=figMuted] at (6.95,4.08)
        {Calibeating};
    \draw[figRule, line width=0.55pt] (0,3.72) -- (7.25,3.72);

    \node[draw=figRule, fill=figPaper, rounded corners=1.3pt,
        align=center, text width=1.72cm, minimum height=0.72cm]
        at (1.35,3.20) {Definitions\\\& notation};
    \node[draw=figRule, fill=figPaper, rounded corners=1.3pt,
        align=center, text width=1.72cm, minimum height=0.72cm]
        at (3.62,3.20) {Problem\\setting};
    \node[draw=figRule, fill=figPaper, rounded corners=1.3pt,
        align=center, text width=1.72cm, minimum height=0.72cm]
        at (5.89,3.20) {Assumptions};

    \draw[figTeal, line width=2.0pt] (0.42,1.08) -- (0.42,2.57);
    \node[anchor=north west, text=figTeal, font=\tiny\bfseries]
        at (0.68,2.56) {TARGET THEOREM};
    \node[align=left, anchor=west, text width=5.85cm, font=\small,
        text=figInk] at (0.68,1.72) {
        \textbf{Proper loss + concave regret bound $\alpha$}\\[2pt]
        \ensuremath{\Rightarrow}\quad
        \textbf{\ensuremath{|Q|\,\alpha(T/|Q|)}-calibeating algorithm}
    };

    % Offline sandbox: several cited-paper graphs form the local library.
    \draw[draw=figRule, line width=0.7pt, fill=white, rounded corners=1.5pt]
        (7.85,0.65) rectangle (13.75,4.45);
    \draw[figBlue, line width=1.35pt] (7.93,4.45) -- (13.67,4.45);
    \node[font=\small\bfseries, text=figInk] at (10.58,4.08) {Sandbox};
    \node[draw=figBlue, fill=figBlueLight, rounded corners=2pt,
        text=figBlue, font=\tiny\bfseries, inner xsep=4pt, inner ysep=1.5pt]
        at (12.92,4.08) {OFFLINE};
    \draw[figRule, line width=0.55pt] (7.85,3.72) -- (13.75,3.72);
    \node[font=\scriptsize\bfseries, text=figInk] at (10.80,3.42)
        {Cited prior work};

    % Three paper silhouettes are illustrative, not a fixed corpus size.
    \foreach \leftedge in {8.18,10.01,11.84} {
        \draw[draw=figRule, fill=white, line width=0.55pt, line join=round]
            (\leftedge,1.68) -- ++(1.56,0) -- ++(0,1.14)
            -- ++(-0.23,0.23) -- ++(-1.33,0) -- cycle;
        \draw[draw=figRule, fill=figPaper, line width=0.5pt]
            (\leftedge+1.33,3.05) -- ++(0,-0.23) -- ++(0.23,0) -- cycle;
    }

    \node[dagnode] (a0) at (8.43,2.34) {};
    \node[dagnode] (a1) at (8.95,2.67) {};
    \node[dagnode] (a2) at (8.95,2.01) {};
    \node[dagnode] (a3) at (9.47,2.34) {};
    \draw[dagflow] (a0) -- (a1);
    \draw[dagflow] (a0) -- (a2);
    \draw[dagflow] (a1) -- (a3);
    \draw[dagflow] (a2) -- (a3);

    \node[dagnode] (b0) at (10.28,2.67) {};
    \node[dagnode] (b1) at (10.28,2.01) {};
    \node[dagnode] (b2) at (10.80,2.34) {};
    \node[dagnode] (b3) at (11.32,2.34) {};
    \draw[dagflow] (b0) -- (b2);
    \draw[dagflow] (b1) -- (b2);
    \draw[dagflow] (b2) -- (b3);

    \node[dagnode] (c0) at (12.11,2.34) {};
    \node[dagnode] (c1) at (12.63,2.67) {};
    \node[dagnode] (c2) at (12.63,2.01) {};
    \node[dagnode] (c3) at (13.15,2.67) {};
    \node[dagnode] (c4) at (13.15,2.01) {};
    \draw[dagflow] (c0) -- (c1);
    \draw[dagflow] (c0) -- (c2);
    \draw[dagflow] (c1) -- (c3);
    \draw[dagflow] (c2) -- (c4);
    \node[text=figBlue, font=\scriptsize\bfseries] at (10.80,1.23)
        {Local Prover Graph};

    % A single compact exclusion cue replaces the former dense table.
    \draw[figCoral, line width=0.7pt, rounded corners=0.7pt]
        (8.20,0.20) rectangle (8.48,0.44);
    \draw[figCoral, line width=0.7pt]
        (8.26,0.44) arc[start angle=180,end angle=0,radius=0.08];
    \node[anchor=west, text=figCoral, font=\tiny\bfseries]
        at (8.62,0.32) {SOURCE PAPER + ITS LEMMAS/PROOFS UNAVAILABLE};

    % Required output.
    \draw[draw=figTeal, line width=0.75pt, fill=figTealLight,
        rounded corners=1.5pt] (14.30,1.42) rectangle (16.55,3.66);
    \draw[figTeal, line width=0.65pt, fill=white]
        (15.11,2.90) rectangle (15.74,3.45);
    \draw[figTeal, line width=0.45pt] (15.22,3.27) -- (15.62,3.27);
    \draw[figTeal, line width=0.45pt] (15.22,3.14) -- (15.58,3.14);
    \draw[figTeal, line width=0.85pt]
        (15.57,2.98) -- (15.68,2.88) -- (15.88,3.10);
    \node[text=figTeal, font=\tiny\bfseries] at (15.43,2.58) {RETURN};
    \node[align=center, text=figInk, font=\scriptsize\bfseries]
        at (15.43,2.02) {One rigorous\\natural-language\\proof};

    % The challenge and sandbox are combined inputs to proof discovery.
    \node[font=\Large\boldmath, text=figMuted, inner sep=0pt] at (7.55,2.55) {$+$};
    \draw[flow] (13.85,2.55) -- (14.20,2.55);
\end{tikzpicture}

%% file: figures/agent_workflows.tex
% Editable vector workflow overview following the supplied four-panel reference.
\begingroup%
\definecolor{wfInk}{HTML}{25333F}%
\definecolor{wfIcon}{HTML}{526D7E}%
\definecolor{wfBorder}{HTML}{A8B7BE}%
\definecolor{wfPanel}{HTML}{FCFDFD}%
\definecolor{wfReturn}{HTML}{B54B40}%
\definecolor{wfGreen}{HTML}{4D9189}%
\definecolor{wfBlueFill}{HTML}{E8F0F5}%
\begin{tikzpicture}[
    x=0.01cm, y=-0.01cm, text=wfInk,
    font=\fontsize{8}{8.5}\selectfont,
    panel/.style={draw=wfBorder,fill=wfPanel,rounded corners=2.2pt,line width=0.6pt},
    card/.style={draw=wfBorder,fill=white,rounded corners=2pt,line width=0.6pt},
    icon/.style={draw=wfIcon,line width=0.6pt,line cap=round,line join=round},
    flow/.style={-{Latex[length=1.7mm,width=1.45mm]},draw=wfIcon,line width=0.85pt},
    dagflow/.style={-{Latex[length=1mm,width=0.8mm]},draw=wfIcon,
        line width=0.65pt,shorten >=3pt},
    feedback/.style={-{Latex[length=1.8mm,width=1.5mm]},draw=wfReturn,line width=0.85pt},
    heading/.style={anchor=north,align=center,inner sep=0pt,
        font=\fontsize{8.5}{9}\selectfont\bfseries},
    description/.style={anchor=north,align=center,inner sep=0pt,
        font=\fontsize{8.2}{8.8}\selectfont\itshape},
    graphnode/.style={draw=wfIcon,fill=white,line width=0.7pt},
    pics/document/.style={code={
        \draw[icon,fill=white] (-20,-28) -- (7,-28) -- (21,-14)
            -- (21,28) -- (-20,28) -- cycle;
        \draw[icon] (7,-28) -- (7,-14) -- (21,-14);
        \draw[icon,line width=0.65pt]
            (-12,-10) -- (1,-10) (-12,-2) -- (13,-2)
            (-12,6) -- (13,6) (-12,14) -- (6,14);
    }},
    pics/prover/.style={code={
        \pic {document};
        \draw[icon,fill=wfBlueFill] (4,27) -- (9,15) -- (32,-8)
            -- (38,-2) -- (15,21) -- cycle;
        \draw[icon,line width=0.55pt] (9,15) -- (15,21) (29,-5) -- (35,1);
    }},
    pics/verifier/.style={code={
        \draw[icon,fill=white] (-6,-7) circle (20);
        \draw[icon,line width=0.6pt] (-18,-9)
            .. controls (-18,-18) and (-11,-22) .. (-4,-21);
        \draw[icon,line width=2.5pt] (9,9) -- (27,27);
    }},
    pics/decider/.style={code={
        \draw[icon,line width=1.1pt] (0,-28) -- (0,27)
            (-23,-20) -- (23,-20) (-13,28) -- (13,28);
        \fill[wfIcon] (0,-28) circle (3);
        \draw[icon,fill=wfBlueFill] (-21,-17) -- (-31,7) -- (-11,7) -- cycle;
        \draw[icon,fill=wfBlueFill] (21,-17) -- (11,7) -- (31,7) -- cycle;
        \draw[icon] (-31,7) .. controls (-28,17) and (-14,17) .. (-11,7)
            (11,7) .. controls (14,17) and (28,17) .. (31,7);
    }},
    pics/decomposer/.style={code={
        \draw[icon] (0,-25) -- (-28,24) (0,-25) -- (28,24);
        \foreach \gx/\gy in {0/-25,-28/24,28/24}
            \draw[graphnode] (\gx,\gy) circle (9);
    }},
    pics/subproofs/.style={code={
        \begin{scope}[shift={(-20,-11)},scale=0.83]\pic {document};\end{scope}
        \begin{scope}[shift={(-5,-5)},scale=0.83]\pic {document};\end{scope}
        \begin{scope}[shift={(12,4)},scale=0.83]\pic {document};\end{scope}
    }},
    pics/recombine/.style={code={
        \draw[icon,fill=wfBlueFill]
            (-28,4) -- (-18,4) .. controls (-23,-8) and (-6,-8) .. (-10,4)
            -- (0,4) -- (0,14) .. controls (12,9) and (12,26) .. (0,22)
            -- (0,32) -- (-28,32) -- (-28,22)
            .. controls (-40,27) and (-40,10) .. (-28,14) -- cycle;
        \draw[icon,fill=white]
            (5,-29) -- (16,-29) .. controls (12,-17) and (29,-17) .. (24,-29)
            -- (35,-29) -- (35,-1) -- (25,-1)
            .. controls (30,11) and (13,11) .. (17,-1)
            -- (5,-1) -- (5,-11)
            .. controls (-7,-7) and (-7,-24) .. (5,-19) -- cycle;
    }},
    pics/planner/.style={code={
        \draw[icon,fill=wfBlueFill] (-9,18) -- (-10,10)
            .. controls (-12,1) and (-20,-2) .. (-20,-14)
            .. controls (-20,-40) and (20,-40) .. (20,-14)
            .. controls (20,-2) and (12,1) .. (10,10)
            -- (9,18) -- cycle;
        \draw[icon] (-9,18) -- (-9,26) -- (9,26) -- (9,18)
            (-7,27) .. controls (-5,35) and (5,35) .. (7,27);
        \draw[icon] (0,-46) -- (0,-39) (-31,-14) -- (-25,-14)
            (25,-14) -- (31,-14) (-24,-38) -- (-19,-32)
            (19,-32) -- (24,-38);
    }},
    pics/finalproof/.style={code={
        \pic {document};
        \draw[fill=wfGreen,draw=white,line width=0.9pt] (22,25) circle (12);
        \draw[draw=white,line width=1pt,line cap=round,line join=round]
            (16,25) -- (20,29) -- (28,21);
    }},
    pics/shallowtree/.style={code={
        \draw[icon] (0,-29) -- (-52,24) (0,-29) -- (-10,24);
        \draw[icon,dashed] (0,-29) -- (47,24);
        \foreach \gx/\gy in {0/-29,-52/24,-10/24,47/24}
            \draw[graphnode,rounded corners=1pt,fill=wfBlueFill!45!white]
                ({\gx-9},{\gy-9}) rectangle ({\gx+9},{\gy+9});
        \node[inner sep=0pt] at (20,25) {$\cdots$};
    }},
    pics/searchtree/.style={code={
        \draw[icon,line width=0.65pt]
            (0,-43) -- (-47,-9) (0,-43) -- (44,-9)
            (-47,-9) -- (-70,25) (-47,-9) -- (-28,25)
            (-28,25) -- (-45,48) (-28,25) -- (-11,48)
            (44,-9) -- (21,25) (44,-9) -- (66,25)
            (66,25) -- (49,48) (66,25) -- (82,48);
        \draw[icon,line width=0.65pt,dashed] (-70,25) -- (-81,54)
            (21,25) -- (12,54) (49,48) -- (45,60);
        \foreach \gx/\gy in {0/-43,-47/-9,44/-9,-70/25,-28/25,21/25,66/25,-45/48,-11/48,49/48,82/48}
            \draw[graphnode,fill=white] (\gx,\gy) circle (7);
        \node[inner sep=0pt] at (0,-5) {$\cdots$};
        \node[inner sep=0pt] at (-68,62) {$\cdots$};
        \node[inner sep=0pt] at (46,66) {$\cdots$};
    }},
    pics/lemmadag/.style={code={
        \draw[dagflow] (0,-42) -- (-39,2);
        \draw[dagflow] (0,-42) -- (31,2);
        \draw[dagflow] (-39,2) -- (-29,50);
        \draw[dagflow] (31,2) -- (3,42);
        \draw[dagflow] (31,2) -- (68,52);
        \draw[dagflow] (-29,50) -- (3,42);
        \foreach \gx/\gy in {0/-42,-39/2,31/2,-29/50,3/42,68/52}
            \draw[graphnode,fill=wfBlueFill!35!white] (\gx,\gy) circle (10);
        \node[inner sep=0pt] at (76,1) {$\cdots$};
        \node[inner sep=0pt] at (36,59) {$\cdots$};
    }}
]
    \path[use as bounding box] (0,0) rectangle (2048,966);
    % Positions and dimensions follow the supplied 2048 x 966 reference.
    % Text is explicitly sized and line-broken, rather than paragraph-wrapped.
    \newcommand{\wftext}[5]{%
        \node[anchor=north,inner sep=0pt,align=center] at (#1,#2)
            {\resizebox*{!}{#3}{#4\shortstack{#5}}};
    }
    \newcommand{\wflefttext}[5]{%
        \node[anchor=north west,inner sep=0pt,align=left] at (#1,#2)
            {\resizebox*{!}{#3}{#4 #5}};
    }
    \newcommand{\wfframe}[4]{%
        \draw[card] (#1,#2) rectangle (#3,#4);
    }
    \newcommand{\wffinalframe}[4]{%
        \draw[card,draw=wfGreen,fill=wfGreen!3!white,line width=0.75pt]
            (#1,#2) rectangle (#3,#4);
    }
    \draw[panel] (4,14) rectangle (1011,463);
    \draw[panel] (1029,14) rectangle (2040,463);
    \draw[panel] (4,487) rectangle (1011,951);
    \draw[panel] (1029,487) rectangle (2040,951);

    % (a) Role-based discussion.
    \wflefttext{40}{32}{0.36cm}{\bfseries}{(a) Role-based discussion}
    \wflefttext{40}{79}{0.24cm}{}{Iterative proof writing with verification and decision.}
    \wfframe{28}{151}{204}{334}
    \wfframe{251}{151}{473}{334}
    \wfframe{522}{151}{744}{334}
    \wffinalframe{834}{151}{990}{334}
    \wftext{116}{173}{0.21cm}{\bfseries}{Prover}
    \wftext{362}{173}{0.21cm}{\bfseries}{Verifier}
    \wftext{633}{173}{0.21cm}{\bfseries}{Decider}
    \wftext{912}{172}{0.44cm}{\bfseries}{Final\\proof}
    \pic[scale=0.82] at (114,239) {prover};
    \pic[scale=0.95] at (361,239) {verifier};
    \pic[scale=0.95] at (633,239) {decider};
    \pic at (910,271) {finalproof};
    \wftext{116}{277}{0.40cm}{\itshape}{Generate\\proof draft}
    \wftext{362}{277}{0.40cm}{\itshape}{Check correctness\\and completeness}
    \wftext{633}{277}{0.40cm}{\itshape}{Decide to accept\\or request revision}
    \draw[flow] (204,239) -- (250,239);
    \draw[flow] (473,239) -- (521,239);
    \draw[flow] (744,239) -- (833,239);
    \wftext{788}{209}{0.20cm}{}{pass}
    \draw[feedback] (639,335) -- (639,394) -- (116,394) -- (116,336);
    \wftext{418}{412}{0.21cm}{\color{wfReturn}}{Continue with shared bounded history}

    % (b) Root-only decomposition.
    \wflefttext{1064}{32}{0.36cm}{\bfseries}{(b) Root-only decomposition}
    \wflefttext{1064}{79}{0.24cm}{}{Decompose the root and prove subgoals.}
    \wfframe{1052}{151}{1233}{335}
    \wfframe{1262}{151}{1440}{335}
    \wfframe{1472}{151}{1654}{335}
    \wfframe{1672}{151}{1841}{335}
    \wffinalframe{1909}{151}{2023}{335}
    \wftext{1142}{174}{0.21cm}{\bfseries}{Decomposer}
    \wftext{1351}{164}{0.45cm}{\bfseries}{Depth-1\\AND--OR tree}
    \wftext{1563}{174}{0.21cm}{\bfseries}{Prove subgoals}
    \wftext{1757}{174}{0.21cm}{\bfseries}{Recombine}
    \wftext{1966}{172}{0.44cm}{\bfseries}{Final\\proof}
    \pic[scale=0.9] at (1142,243) {decomposer};
    \pic[scale=0.95] at (1564,244) {subproofs};
    \pic[scale=0.9] at (1757,245) {recombine};
    \pic at (1963,271) {finalproof};
    % Depth-1 tree: one layer of subgoals, with alternatives shown schematically.
    \draw[icon] (1351,237) -- (1331,283);
    \draw[icon,dashed] (1351,237) -- (1290,283) (1351,237) -- (1395,283);
    \foreach \gx/\gy in {1351/237,1290/283,1331/283,1395/283}
        \draw[graphnode,rounded corners=0.7pt,fill=wfBlueFill!40!white]
            ({\gx-9},{\gy-9}) rectangle ({\gx+9},{\gy+9});
    \wftext{1362}{276}{0.045cm}{}{$\cdots$}
    \wftext{1418}{286}{0.045cm}{\color{wfBorder}}{$\cdots$}
    \wftext{1142}{285}{0.38cm}{\itshape}{Decompose the goal\\into subgoals}
    \wftext{1563}{285}{0.38cm}{\itshape}{Prove each\\planned subgoal}
    \wftext{1757}{292}{0.19cm}{\itshape}{Combine subproofs}
    \draw[flow] (1233,239) -- (1261,239);
    \draw[flow] (1440,239) -- (1471,239);
    \draw[flow] (1654,239) -- (1671,239);
    \draw[flow] (1841,239) -- (1908,239);
    \wftext{1874}{209}{0.20cm}{}{pass}
    \draw[feedback] (1754,336) -- (1754,395) -- (1142,395) -- (1142,337);
    \wftext{1450}{412}{0.21cm}{\color{wfReturn}}{Retry with a new decomposition}

    % (c) Decomposition with MCTS; the tree card is deliberately wider.
    \wflefttext{40}{507}{0.36cm}{\bfseries}{(c) Decomposition with MCTS}
    \wflefttext{40}{554}{0.25cm}{\itshape}{Search over decompositions using Monte Carlo Tree Search.}
    \wfframe{20}{624}{185}{814}
    \wfframe{211}{620}{423}{818}
    \wfframe{450}{624}{634}{814}
    \wfframe{647}{624}{818}{814}
    \wffinalframe{879}{624}{993}{814}
    \wftext{102}{648}{0.21cm}{\bfseries}{Decomposer}
    \wftext{317}{638}{0.17cm}{\bfseries}{Full AND--OR tree}
    \wftext{542}{648}{0.21cm}{\bfseries}{Prove subgoals}
    \wftext{732}{648}{0.21cm}{\bfseries}{Recombine}
    \wftext{936}{646}{0.44cm}{\bfseries}{Final\\proof}
    \pic[scale=0.9] at (102,715) {decomposer};
    \pic[scale=1.1] at (317,728) {searchtree};
    \pic[scale=0.95] at (541,718) {subproofs};
    \pic[scale=0.9] at (739,720) {recombine};
    \pic at (934,745) {finalproof};
    \wftext{102}{759}{0.40cm}{\itshape}{Propose\\decomposition}
    \wftext{542}{762}{0.39cm}{\itshape}{Prove selected\\subgoals}
    \wftext{732}{766}{0.19cm}{\itshape}{Combine subproofs}
    \draw[flow] (185,716) -- (210,716);
    \draw[flow] (423,716) -- (449,716);
    \draw[flow] (634,716) -- (646,716);
    \draw[flow] (818,716) -- (878,716);
    \wftext{848}{686}{0.20cm}{}{pass}
    \draw[feedback] (734,815) -- (734,874) -- (99,874) -- (99,818);
    \wftext{420}{893}{0.21cm}{\color{wfReturn}}{Select promising nodes and expand the search}

    % (d) Agentic planning.
    \wflefttext{1064}{507}{0.36cm}{\bfseries}{(d) Agentic planning}
    \wflefttext{1064}{554}{0.25cm}{\itshape}{Plan the proof via a lemma dependency graph and iteratively refine it.}
    \wfframe{1047}{624}{1228}{814}
    \wfframe{1267}{624}{1448}{814}
    \wfframe{1483}{624}{1680}{814}
    \wfframe{1700}{624}{1855}{814}
    \wffinalframe{1919}{624}{2030}{814}
    \wftext{1138}{648}{0.21cm}{\bfseries}{Planner}
    \wftext{1358}{648}{0.21cm}{\bfseries}{Lemma DAG}
    \wftext{1581}{648}{0.21cm}{\bfseries}{Prove subgoals}
    \wftext{1778}{648}{0.21cm}{\bfseries}{Recombine}
    \wftext{1975}{646}{0.44cm}{\bfseries}{Final\\proof}
    \pic[scale=0.82] at (1138,720) {planner};
    \pic at (1340,733) {lemmadag};
    \pic[scale=0.95] at (1577,718) {subproofs};
    \pic[scale=0.9] at (1785,720) {recombine};
    \pic at (1973,745) {finalproof};
    \wftext{1138}{763}{0.39cm}{\itshape}{Plan useful lemmas\\and proof strategy}
    \wftext{1581}{763}{0.39cm}{\itshape}{Prove subgoals\\according to the plan}
    \wftext{1778}{763}{0.39cm}{\itshape}{Assemble into\\a final proof}
    \draw[flow] (1228,716) -- (1266,716);
    \draw[flow] (1448,716) -- (1482,716);
    \draw[flow] (1680,716) -- (1699,716);
    \draw[flow] (1855,716) -- (1918,716);
    \wftext{1888}{686}{0.20cm}{}{pass}
    \draw[feedback] (1774,815) -- (1774,874) -- (1134,874) -- (1134,818);
    \wftext{1456}{893}{0.21cm}{\color{wfReturn}}{Revise lemma DAG and plan}
\end{tikzpicture}%
\endgroup%

%% file: sections/illustrative_evaluation.tex
% Keep paragraph spacing local to Section 4; figure spacing is shared.
\begingroup
\setlength{\parskip}{4pt}

\section{Model and Agent Comparisons}\label{sec:evaluation}

We evaluate TCSAlgBench along two axes: model capability under common inference settings and agent design under matched model-call opportunities. After describing the experimental setup, we compare model configurations under direct inference and 10-round discussion, then compare four agent workflows using a fixed backbone model. We conclude the section with diagnostic analyses.

\subsection{Experimental Setup}\label{sec:experimental-setup}

\paragraph{Models.}
We evaluate Opus~4.8 and GPT-5.6 Sol at high, xhigh, and max effort, and GPT-5.5 and Fable~5 at high and xhigh. Each configuration is evaluated under direct inference and a simple 10-round prover--verifier discussion workflow. Direct inference uses the common proof prompt once, without critique or revision; discussion iteratively refines the proof using verifier feedback. We use ten independent runs for direct inference and five for discussion. All models are called through Amazon Bedrock with a 128K-token per-call output cap. We excluded Fable~5 max because of output-token access limits.

\paragraph{Agent designs.}
An agent workflow specifies how proving, verification, decomposition, and planning actions are coordinated in response to intermediate results. Using GPT-5.5 xhigh, we compare four workflows representing recurring proof-agent mechanisms (Figure~\ref{fig:agent-workflows}). Iterative generation and critique motivate \emph{role-based discussion}~\citep{fengEtAl26,anYPZ26,schmittEtAl26proofcouncil}; proof sketches and structured subgoals motivate \emph{root-only decomposition}~\citep{jiangEtAl23dsp,zhangEtAl25cumulative,varamballyEtAl25hilbert}; AND--OR search and value-guided exploration motivate \emph{decomposition with MCTS}~\citep{lampleEtAl22htps,xinEtAl25deepseek,tsoukalasEtAl26alphaproofnexus}, with implementation details in Appendix~\ref{app:agent-workflows}; and replanning with revisable lemma DAGs motivates \emph{agentic planning}~\citep{anYPZ26,wuEtAl26starpolya,chungEtAl26goedelarchitect}. These workflows abstract common mechanisms and implement them using shared prover and verifier prompts, enabling a fair comparison of agent designs with a fixed backbone model and matched model-call opportunities. We use GPT-5.5 xhigh because its earlier knowledge cutoff leaves a larger subset of challenges published after the cutoff.

Each workflow uses five independent runs with up to 20 outer iterations per run, matched model-call opportunities and per-call limits, and the same external verifier. This experiment therefore differs from the model comparison in both its discussion protocol and its search budget. Appendix~\ref{sec:evaluation_protocol} details the workflows, budgets, and scoring rules. Our GPT-5.5 xhigh agent with agentic planning achieves higher five-run coverage than GPT-5.6 Sol max with discussion and repeated sampling, demonstrating the effectiveness of our agent implementation.

\subsection{The Benchmark Separates Model Configurations}\label{sec:model-results}

Fable~5 leads direct inference: xhigh has the highest seed-1 acceptance (6.8\%), while high has the highest ten-run coverage (14.6\%). Under 10-round discussion, GPT-5.6 Sol max achieves the highest seed-1 acceptance (18.8\%) and five-run coverage (23.6\%). This change in ordering shows that model comparisons depend on how inference is organized. Even the strongest configuration leaves more than three-quarters of the benchmark uncovered after five discussion runs (Figure~\ref{fig:model-comparison}).

\begin{figure}[!ht]
    \centering
    \resizebox{0.9\textwidth}{!}{%
        \input{figures/model_comparison}%
    }
    \par\nointerlineskip
    \caption{Verifier acceptance on all 398 challenges under the three-voter majority rule. Blue bars show seed~1; red bars show coverage over 10 direct-inference runs (a) or five 10-round discussion runs (b). Labels give percentages. Rows are grouped by model family and ordered within each family by coverage; Tables~\ref{tab:direct-inference-baseline} and~\ref{tab:all-topic} give detailed results.}
    \label{fig:model-comparison}
\end{figure}
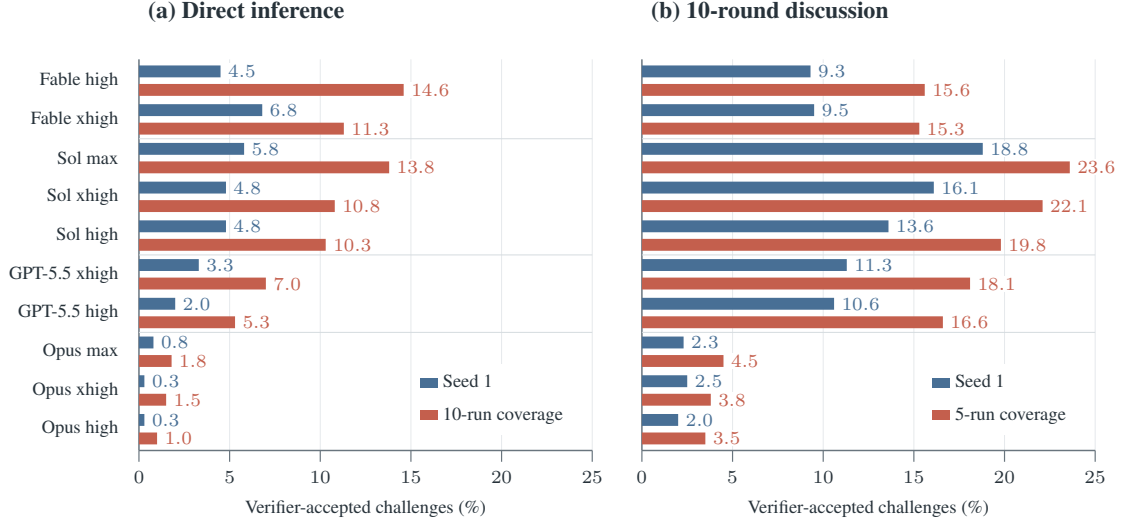

\paragraph{Discussion and repeated sampling.}
We compare ten independent direct-inference runs with one 10-round discussion run, allocating proof-generation attempts to independent sampling or successive revisions. One discussion run outperforms ten-run direct coverage for all eight GPT and Opus configurations: Sol max reaches 18.8\% versus 13.8\%, and GPT-5.5 xhigh reaches 11.3\% versus 7.0\%. Fable~5 is the sole exception: ten-run direct coverage reaches 14.6\% and 11.3\% for high and xhigh, respectively, compared with 9.3\% and 9.5\% for one discussion run (Tables~\ref{tab:direct-inference-baseline} and~\ref{tab:all-topic}). These results suggest that feedback-guided revision can be more productive than repeatedly starting from scratch, while the effective allocation of inference attempts depends on the model. Independent restarts remain useful after multi-round discussion. Five discussion runs increase coverage from 18.8\% to 23.6\% for Sol max and from 11.3\% to 18.1\% for GPT-5.5 xhigh. Thus, a single refinement trajectory leaves substantial additional coverage accessible through other seeds. Together, these observations motivate combining prover--verifier feedback with multiple independent starts.

\paragraph{Effects of reasoning effort.}
Higher effort does not uniformly improve coverage. Fable~5 high covers 58 challenges across ten direct-inference runs, versus 45 for xhigh, despite averaging 46K rather than 62K output tokens per call. Under five-run discussion, Sol xhigh and max share 78 accepted challenges, with 10 unique to xhigh and 16 unique to max (Table~\ref{tab:effort-complementarity}). These complementary coverage sets suggest that varying effort can expose alternative successful approaches; simply choosing the highest effort does not subsume the results of lower settings.

\subsection{Agent Design Changes Effectiveness and Cost}\label{sec:agent-results}

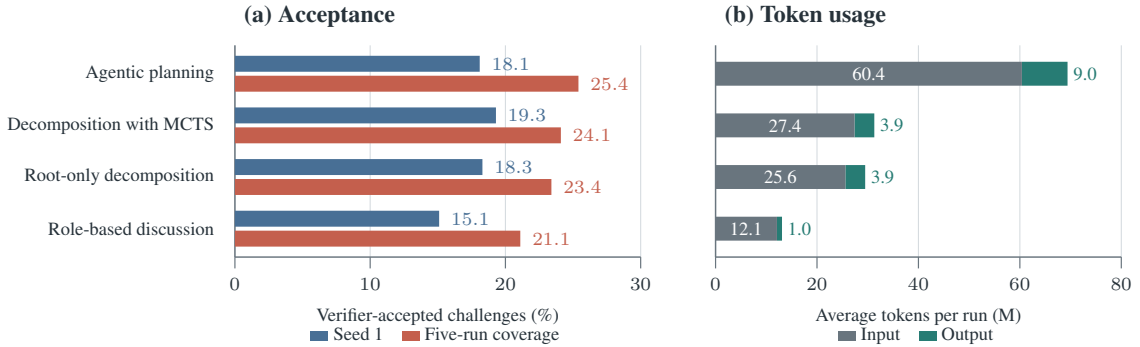
\begin{figure}[!ht]
    \centering
    \input{figures/agent_comparison}
    \par\nointerlineskip
    \caption{Call-matched agent-design comparison using GPT-5.5 xhigh. (a) Seed-1 acceptance and five-run verifier-accepted coverage on all 398 challenges. (b) Input and output tokens averaged over 10 randomly selected challenges and five independent seeds; M denotes millions. Rows follow five-run coverage. Workflows share call opportunities and per-call generation limits, but token usage differs. Table~\ref{tab:preliminary-agent-ablation} gives exact counts and budget details.}
    \label{fig:agent-comparison}
\end{figure}

We compare four agent workflows using GPT-5.5 xhigh under matched model-call opportunities (Figure~\ref{fig:agent-comparison}). Only final assembled proofs contribute to acceptance.

\paragraph{Decomposition and search.}
Root-only decomposition increases seed-1 acceptance from 15.1\% to 18.3\% and five-run coverage from 21.1\% to 23.4\% relative to discussion without decomposition. These gains suggest that organizing a proof around explicit intermediate claims helps the model make progress. The quality of the decomposition matters: its subgoals must be easier to establish and jointly sufficient to prove the target theorem. Identifying such intermediate claims is itself a substantive reasoning task, making decomposition an important part of proof discovery. MCTS achieves the highest seed-1 acceptance (19.3\%), but its five-run coverage exceeds root-only decomposition by only three challenges (96 versus 93). One possible bottleneck is the quality of the LLM-generated scores used to guide search: the model may not be sufficiently calibrated to assess promising subproblems in these proof tasks.

\paragraph{Adaptive planning improves coverage at higher token cost.}
Agentic planning achieves higher five-run coverage than MCTS (25.4\% versus 24.1\%), although its seed-1 acceptance is lower (18.1\% versus 19.3\%). The planner proposes decompositions globally, considering how the subgoals and their dependencies fit together to prove the target theorem. This global view may produce more coherent decompositions and help explain its broader coverage. In the token sample, planning averages 60.4M input and 9.0M output tokens per run, versus 27.4M and 3.9M for MCTS. The global plan provides more proof targets to explore from the outset, which may help explain the higher token consumption under matched call opportunities.

\subsection{Diagnostic Analyses and Broader Uses}\label{sec:robustness-results}

\paragraph{Topic-specific performance.}
Differential-privacy challenges seem to be a challenging area for LLMs: nine of ten configurations cover at most one of the 33 sampled tasks after five 10-round discussion runs (Table~\ref{tab:all-topic}). Sol max covers two (6.1\%), versus 23.6\% average (Figure~\ref{fig:topic-coverage}).

\begin{figure}[!ht]
    \centering
    \input{figures/topic_coverage}
    \par\nointerlineskip
    \caption{Topic performance and benchmark composition. (a) Five-run verifier-accepted coverage for GPT-5.6 Sol max after 10-round prover--verifier discussion. Labels show within-topic coverage and accepted/total counts; the dashed line marks overall coverage (94/398, 23.6\%). (b) Topic shares of all 398 challenges. Colors and abbreviations match across panels.}
    \label{fig:topic-coverage}
\end{figure}
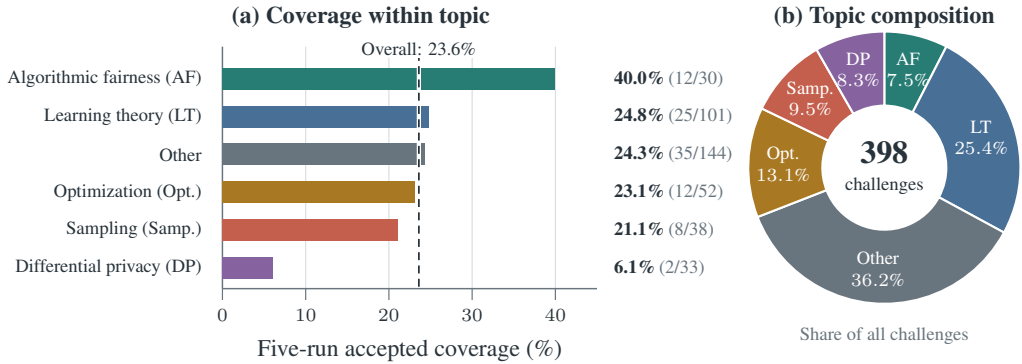

\paragraph{Verifier sensitivity.}
Re-scoring identical proofs with Opus~4.8 high effort checks the stability of the comparisons. The eight shared model configurations retain their five-run coverage ordering, with count changes of $-6$ to $+2$ (Table~\ref{tab:verifier-sensitivity}). Under the GPT-5.5 three-voter majority rule versus the Opus~4.8 verifier, five-run coverage is 101 versus 103 challenges for planning, 93 versus 96 for root-only decomposition, 84 versus 72 for discussion, and 96 versus 84 for MCTS. Both verifiers support our conclusion that MCTS offers no clear advantage over root-only decomposition.

\paragraph{Source-date diagnostics.}\label{sec:cutoff-results}
Figure~\ref{fig:cutoff-analysis} partitions the 10-round discussion results by each source paper's first arXiv date relative to the model family's assumed cutoff. Post-cutoff five-run coverage is higher for seven of ten configurations, with post-minus-pre differences of $-4.0$ to $+4.3$ percentage points (Table~\ref{tab:scale-cutoff}). There is no consistent pre-cutoff advantage.

\begin{figure}[!ht]
    \centering
    \input{figures/cutoff_analysis}
    \par\nointerlineskip
    \caption{Five-run verifier-accepted coverage before and after assumed knowledge cutoffs. Each panel shows its cutoff date and pre-/post-cutoff sample sizes (total $n=398$). Axes share the same scale; comparisons are within configurations because cutoff dates differ across families.}
    \label{fig:cutoff-analysis}
\end{figure}
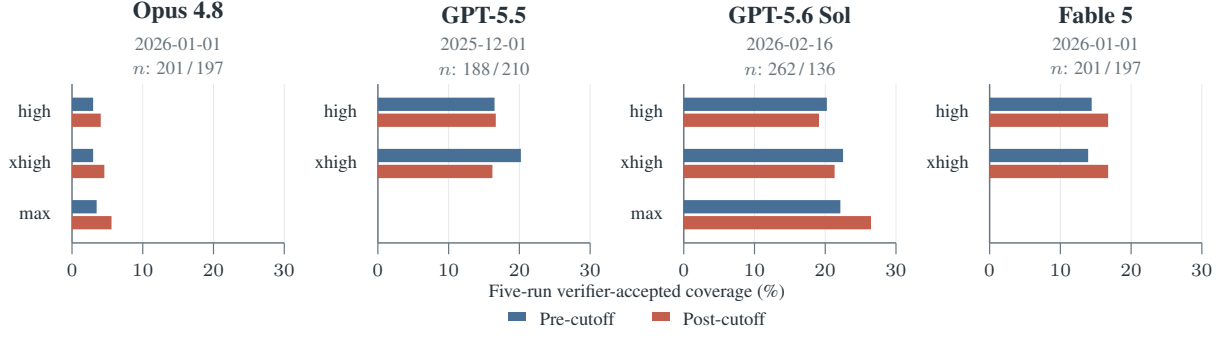

\paragraph{Diagnosing model behavior.}
Beyond evaluating mathematical proof performance, TCSAlgBench may support broader studies of model behavior. For example, comparing challenge-level overlap across effort settings could help investigate how additional inference effort changes coverage, as suggested by the complementary Sol xhigh and max results. Fable~5 xhigh's higher direct-inference token use and smaller seed-1 gain from discussion than Sol max could motivate studies of how model-specific prompting or internal refinement interacts with external critique. Source-date partitions and fresh challenge batches could also support investigations of possible training exposure. These examples illustrate potential diagnostic uses of the benchmark and directions for future study.

\endgroup

%% file: figures/model_comparison.tex
\begin{tikzpicture}
    \pgfplotsset{
        model comparison bars/.style={
            bench axis,
            scale only axis,
            width=0.365\textwidth,
            height=5.2cm,
            xlabel={Verifier-accepted challenges (\%)},
            xlabel style={at={(0.5,-0.105)},anchor=north,font=\scriptsize},
            xmin=0, xmax=25,
            xtick={0,5,10,15,20,25},
            ytick={0,1,2,3,4,5,6,7,8,9},
            ymin=-0.55, ymax=9.55,
            title style={at={(0,1.03)},anchor=south west},
            grid style={draw=figRule!60, line width=0.35pt},
            xbar,
            bar width=4.5pt,
            point meta=x,
            nodes near coords={\pgfmathprintnumber[fixed,precision=1,zerofill]{\pgfplotspointmeta}},
            every node near coord/.append style={font=\scriptsize,
                anchor=west, xshift=2pt, fill=white, inner sep=0.7pt},
            legend image code/.code={\path[##1] (0cm,-0.06cm) rectangle (0.23cm,0.06cm);},
            legend style={at={(0.96,0.04)},anchor=south east,legend columns=1,
                legend cell align=left,fill=white,inner sep=2pt,row sep=2pt},
        }
    }
    \begin{axis}[
        model comparison bars,
        name=direct,
        title={\textbf{(a) Direct inference}},
        yticklabels={Opus high,Opus xhigh,Opus max,GPT-5.5 high,GPT-5.5 xhigh,Sol high,Sol xhigh,Sol max,Fable xhigh,Fable high},
    ]
        \draw[figRule, line width=0.35pt] (axis cs:0,2.5) -- (axis cs:25,2.5);
        \draw[figRule, line width=0.35pt] (axis cs:0,4.5) -- (axis cs:25,4.5);
        \draw[figRule, line width=0.35pt] (axis cs:0,7.5) -- (axis cs:25,7.5);

        % Separate vertical positions keep both labels legible for small gaps.
        \addplot[fill=figBlue, draw=none, bar shift=3.5pt,
            every node near coord/.append style={text=figBlue}] coordinates {
            (0.3,0) (0.3,1) (0.8,2) (2.0,3) (3.3,4) (4.8,5)
            (4.8,6) (5.8,7) (6.8,8) (4.5,9)
        };
        \addlegendentry{Seed 1}
        \addplot[fill=figCoral, draw=none, bar shift=-3.5pt,
            every node near coord/.append style={text=figCoral}] coordinates {
            (1.0,0) (1.5,1) (1.8,2) (5.3,3) (7.0,4) (10.3,5)
            (10.8,6) (13.8,7) (11.3,8) (14.6,9)
        };
        \addlegendentry{10-run coverage}
    \end{axis}

    \begin{axis}[
        model comparison bars,
        at={(direct.east)},
        anchor=west,
        xshift=0.04\textwidth,
        title={\textbf{(b) 10-round discussion}},
        yticklabel=\empty,
    ]
        \draw[figRule, line width=0.35pt] (axis cs:0,2.5) -- (axis cs:25,2.5);
        \draw[figRule, line width=0.35pt] (axis cs:0,4.5) -- (axis cs:25,4.5);
        \draw[figRule, line width=0.35pt] (axis cs:0,7.5) -- (axis cs:25,7.5);

        \addplot[fill=figBlue, draw=none, bar shift=3.5pt,
            every node near coord/.append style={text=figBlue}] coordinates {
            (2.0,0) (2.5,1) (2.3,2) (10.6,3) (11.3,4) (13.6,5)
            (16.1,6) (18.8,7) (9.5,8) (9.3,9)
        };
        \addlegendentry{Seed 1}
        \addplot[fill=figCoral, draw=none, bar shift=-3.5pt,
            every node near coord/.append style={text=figCoral}] coordinates {
            (3.5,0) (3.8,1) (4.5,2) (16.6,3) (18.1,4) (19.8,5)
            (22.1,6) (23.6,7) (15.3,8) (15.6,9)
        };
        \addlegendentry{5-run coverage}
    \end{axis}
\end{tikzpicture}

%% file: figures/agent_comparison.tex
\begin{tikzpicture}
    \begin{axis}[
        bench axis,
        name=agentAcceptance,
        scale only axis,
        width=0.325\textwidth,
        height=2.8cm,
        xmin=0, xmax=30,
        xtick={0,10,20,30},
        xlabel={Verifier-accepted challenges (\%)},
        xlabel style={font=\scriptsize},
        ytick={0,1,2,3},
        yticklabels={Role-based discussion,Root-only decomposition,Decomposition with MCTS,Agentic planning},
        ymin=-0.55, ymax=3.55,
        xbar,
        bar width=6pt,
        point meta=x,
        nodes near coords={\pgfmathprintnumber[fixed,precision=1,zerofill]{\pgfplotspointmeta}},
        every node near coord/.append style={font=\scriptsize,
            anchor=west, xshift=1pt},
        legend image code/.code={\path[#1] (0cm,-0.06cm) rectangle (0.23cm,0.06cm);},
        legend style={at={(0.5,-0.28)},anchor=north,legend columns=2,
            /tikz/every even column/.append style={column sep=5pt}},
    ]
        % Rows follow five-run coverage; seed 1 is a single execution.
        \addplot[fill=figBlue, draw=none, bar shift=3.75pt,
            every node near coord/.append style={text=figBlue}] coordinates {
            (15.1,0) (18.3,1) (19.3,2) (18.1,3)
        };
        \addlegendentry{Seed 1}
        \addplot[fill=figCoral, draw=none, bar shift=-3.75pt,
            every node near coord/.append style={text=figCoral}] coordinates {
            (21.1,0) (23.4,1) (24.1,2) (25.4,3)
        };
        \addlegendentry{Five-run coverage}
    \end{axis}

    \begin{axis}[
        bench axis,
        name=agentTokens,
        at={(agentAcceptance.east)},
        anchor=west,
        xshift=0.06\textwidth,
        scale only axis,
        width=0.325\textwidth,
        height=2.8cm,
        xmin=0, xmax=80,
        xtick={0,20,40,60,80},
        xlabel={Average tokens per run (M)},
        xlabel style={font=\scriptsize},
        ytick={0,1,2,3},
        yticklabel=\empty,
        ymin=-0.55, ymax=3.55,
        xbar stacked,
        bar width=9pt,
        bar shift=0pt,
        legend image code/.code={\path[#1] (0cm,-0.06cm) rectangle (0.23cm,0.06cm);},
        legend style={at={(0.5,-0.28)},anchor=north,legend columns=2,
            /tikz/every even column/.append style={column sep=5pt}},
    ]
        % Token averages are per challenge--seed execution over 10 randomly
        % selected challenges and five independent seeds.
        \addplot[fill=figMuted, draw=none] coordinates {
            (12.1,0) (25.6,1) (27.4,2) (60.4,3)
        };
        \addlegendentry{Input}
        \addplot[fill=figTeal, draw=none] coordinates {
            (1.0,0) (3.9,1) (3.9,2) (9.0,3)
        };
        \addlegendentry{Output}

        % Input labels sit inside their segments; output labels follow the
        % bar ends so that even the 1.0M segment remains legible.
        \foreach \inputTokens/\outputTokens/\row in {
            12.1/1.0/0,25.6/3.9/1,27.4/3.9/2,60.4/9.0/3} {
            \pgfmathsetmacro{\inputMidpoint}{\inputTokens/2}
            \pgfmathsetmacro{\totalTokens}{\inputTokens+\outputTokens}
            \edef\tokenAnchors{\noexpand\coordinate (agentInput\row)
                at (axis cs:\inputMidpoint,\row);
                \noexpand\coordinate (agentOutput\row)
                at (axis cs:\totalTokens,\row);}
            \tokenAnchors
        }
    \end{axis}

    % Draw labels after the axis so that filled bars cannot cover them.
    \foreach \inputTokens/\outputTokens/\row in {
        12.1/1.0/0,25.6/3.9/1,27.4/3.9/2,60.4/9.0/3} {
        \node[font=\scriptsize, text=white, inner sep=0pt]
            at (agentInput\row) {\inputTokens};
        \node[anchor=west, font=\scriptsize, text=figTeal,
            inner sep=0pt, xshift=2pt]
            at (agentOutput\row) {\outputTokens};
    }

    \node[anchor=south west, font=\small\bfseries, text=figInk]
        at ([yshift=0.12cm]agentAcceptance.north west)
        {(a) Acceptance};
    \node[anchor=south west, font=\small\bfseries, text=figInk]
        at ([yshift=0.12cm]agentTokens.north west)
        {(b) Token usage};
\end{tikzpicture}

%% file: figures/topic_coverage.tex
\begingroup
\colorlet{topicAF}{figTeal}
\colorlet{topicLT}{figBlue}
\colorlet{topicOther}{figMuted}
\colorlet{topicOpt}{figAmber}
\colorlet{topicSamp}{figCoral}
\definecolor{topicDP}{HTML}{86639F}
\begin{tikzpicture}
    \begin{axis}[
        bench axis,
        name=topicRates,
        scale only axis,
        width=0.30\textwidth,
        height=3.2cm,
        xmin=0, xmax=45,
        xtick={0,10,20,30,40},
        xlabel={Five-run accepted coverage (\%)},
        ytick={0,1,2,3,4,5},
        yticklabels={Differential privacy (DP),Sampling (Samp.),Optimization (Opt.),Other,Learning theory (LT),Algorithmic fairness (AF)},
        ymin=-0.55, ymax=5.85,
        xbar,
        bar width=8pt,
        bar shift=0pt,
    ]
        % Within-topic rates: accepted / total, from Table tab:all-topic, Panel B.
        \addplot[fill=topicDP, draw=none] coordinates {(6.1,0)};
        \addplot[fill=topicSamp, draw=none] coordinates {(21.1,1)};
        \addplot[fill=topicOpt, draw=none] coordinates {(23.1,2)};
        \addplot[fill=topicOther, draw=none] coordinates {(24.3,3)};
        \addplot[fill=topicLT, draw=none] coordinates {(24.8,4)};
        \addplot[fill=topicAF, draw=none] coordinates {(40.0,5)};

        % A white underlay keeps the overall reference visible across colored bars.
        \draw[white, line width=1.6pt]
            (axis cs:23.6,-0.5) -- (axis cs:23.6,5.55);
        \draw[figInk, densely dashed, line width=0.65pt]
            (axis cs:23.6,-0.5) -- (axis cs:23.6,5.55);
        \node[anchor=south, font=\scriptsize, text=figInk,
            fill=white, inner sep=1pt]
            at (axis cs:23.6,5.55) {Overall: 23.6\%};

        % One aligned annotation column keeps the labels clear of the bars.
        \node[anchor=west, font=\scriptsize, text=figInk, inner sep=0pt]
            at (axis cs:47,0) {{\bfseries 6.1\%}\,\textcolor{figMuted}{(2/33)}};
        \node[anchor=west, font=\scriptsize, text=figInk, inner sep=0pt]
            at (axis cs:47,1) {{\bfseries 21.1\%}\,\textcolor{figMuted}{(8/38)}};
        \node[anchor=west, font=\scriptsize, text=figInk, inner sep=0pt]
            at (axis cs:47,2) {{\bfseries 23.1\%}\,\textcolor{figMuted}{(12/52)}};
        \node[anchor=west, font=\scriptsize, text=figInk, inner sep=0pt]
            at (axis cs:47,3) {{\bfseries 24.3\%}\,\textcolor{figMuted}{(35/144)}};
        \node[anchor=west, font=\scriptsize, text=figInk, inner sep=0pt]
            at (axis cs:47,4) {{\bfseries 24.8\%}\,\textcolor{figMuted}{(25/101)}};
        \node[anchor=west, font=\scriptsize, text=figInk, inner sep=0pt]
            at (axis cs:47,5) {{\bfseries 40.0\%}\,\textcolor{figMuted}{(12/30)}};
    \end{axis}

    \node[anchor=south west, font=\small\bfseries, text=figInk]
        at ([yshift=0.12cm]topicRates.north west)
        {(a) Coverage within topic};
    \coordinate (topicDonut) at ([xshift=3.8cm]topicRates.east);
    \node[anchor=south, font=\small\bfseries, text=figInk]
        at ([xshift=3.8cm,yshift=0.12cm]topicRates.north east)
        {(b) Topic composition};
    \begin{scope}[shift={(topicDonut)}]
        % Slice areas use all 398 challenges, not the 94 accepted challenges.
        % The order and colors match the bar chart, from its top row downward.
        \foreach \topicCount/\topicColor/\topicLabel
            [remember=\topicEndAngle as \topicStartAngle (initially 90)]
            in {30/topicAF/AF,101/topicLT/LT,144/topicOther/Other,
                52/topicOpt/Opt.,38/topicSamp/Samp.,33/topicDP/DP} {
            \pgfmathsetmacro{\topicEndAngle}{\topicStartAngle-360*(\topicCount/398)}
            \pgfmathsetmacro{\topicMidAngle}{(\topicStartAngle+\topicEndAngle)/2}
            \pgfmathsetmacro{\topicShare}{100*(\topicCount/398)}
            \path[fill=\topicColor, draw=white, line width=0.8pt]
                (\topicStartAngle:1.8)
                arc[start angle=\topicStartAngle,end angle=\topicEndAngle,radius=1.8cm]
                -- (\topicEndAngle:0.82)
                arc[start angle=\topicEndAngle,end angle=\topicStartAngle,radius=0.82cm]
                -- cycle;
            \node[align=center, text=white, inner sep=0pt, font=\scriptsize]
                at (\topicMidAngle:1.34)
                {\topicLabel\\\pgfmathprintnumber[fixed,precision=1,zerofill]{\topicShare}\%};
        }
        \node[align=center, text=figInk, inner sep=0pt]
            at (0,0) {{\large\bfseries 398}\\[-0.3mm]{\scriptsize challenges}};
        \node[anchor=north, font=\scriptsize, text=figMuted]
            at (0,-1.99) {Share of all challenges};
    \end{scope}
\end{tikzpicture}
\endgroup

%% file: figures/cutoff_analysis.tex
\begin{tikzpicture}
    \pgfplotsset{
        cutoff bars/.style={
            bench axis,
            scale only axis,
            width=0.170\textwidth,
            height=2.10cm,
            xmin=0, xmax=30,
            xtick={0,10,20,30},
            ymin=-0.55, ymax=2.55,
            ytick={0,1,2},
            yticklabels={max,xhigh,high},
            grid style={draw=figRule!60, line width=0.35pt},
            xbar,
            bar width=5pt,
        }
    }
    % Exact counts and denominators come from Table tab:scale-cutoff, Panel B.
    % All panels use the same percentage scale and high/xhigh/max row order.
    \begin{axis}[cutoff bars, name=cutoffOpus]
        \addplot[fill=figBlue, draw=none, bar shift=3pt] coordinates {
            ({100*6/201},2) ({100*6/201},1) ({100*7/201},0)
        };
        \addplot[fill=figCoral, draw=none, bar shift=-3pt] coordinates {
            ({100*8/197},2) ({100*9/197},1) ({100*11/197},0)
        };
    \end{axis}
    \begin{axis}[cutoff bars, name=cutoffGPT,
        at={(cutoffOpus.east)}, anchor=west, xshift=0.075\textwidth,
        ytick={1,2}, yticklabels={xhigh,high}]
        \addplot[fill=figBlue, draw=none, bar shift=3pt] coordinates {
            ({100*31/188},2) ({100*38/188},1)
        };
        \addplot[fill=figCoral, draw=none, bar shift=-3pt] coordinates {
            ({100*35/210},2) ({100*34/210},1)
        };
    \end{axis}
    \begin{axis}[cutoff bars, name=cutoffSol,
        at={(cutoffGPT.east)}, anchor=west, xshift=0.075\textwidth]
        \addplot[fill=figBlue, draw=none, bar shift=3pt] coordinates {
            ({100*53/262},2) ({100*59/262},1) ({100*58/262},0)
        };
        \addplot[fill=figCoral, draw=none, bar shift=-3pt] coordinates {
            ({100*26/136},2) ({100*29/136},1) ({100*36/136},0)
        };
    \end{axis}
    \begin{axis}[cutoff bars, name=cutoffFable,
        at={(cutoffSol.east)}, anchor=west, xshift=0.075\textwidth,
        ytick={1,2}, yticklabels={xhigh,high}]
        \addplot[fill=figBlue, draw=none, bar shift=3pt] coordinates {
            ({100*29/201},2) ({100*28/201},1)
        };
        \addplot[fill=figCoral, draw=none, bar shift=-3pt] coordinates {
            ({100*33/197},2) ({100*33/197},1)
        };
    \end{axis}

    \foreach \panel/\family/\cutoffDate/\preCount/\postCount in {
        cutoffOpus/Opus 4.8/2026-01-01/201/197,
        cutoffGPT/GPT-5.5/2025-12-01/188/210,
        cutoffSol/GPT-5.6 Sol/2026-02-16/262/136,
        cutoffFable/Fable 5/2026-01-01/201/197} {
        \node[anchor=south, font=\small\bfseries, text=figInk]
            at ([yshift=0.68cm]\panel.north) {\family};
        \node[anchor=south, font=\scriptsize, text=figMuted]
            at ([yshift=0.33cm]\panel.north) {\cutoffDate};
        \node[anchor=south, font=\scriptsize, text=figMuted]
            at (\panel.north) {$n$: \preCount\,/\,\postCount};
    }
    \coordinate (cutoffMidpoint) at ($(cutoffOpus.south west)!0.5!(cutoffFable.south east)$);
    \node[anchor=north, font=\scriptsize, text=figInk]
        at ([yshift=-0.42cm]cutoffMidpoint) {Five-run verifier-accepted coverage (\%)};
    \node[anchor=north, font=\scriptsize, text=figInk]
        at ([yshift=-0.79cm]cutoffMidpoint) {
            \tikz[baseline=-0.5ex]\fill[figBlue] (0,0) rectangle (0.23,0.11);
            \enspace Pre-cutoff\qquad
            \tikz[baseline=-0.5ex]\fill[figCoral] (0,0) rectangle (0.23,0.11);
            \enspace Post-cutoff
        };
\end{tikzpicture}

%% file: sections/conclusion.tex
% Keep the closing statements on the final main-text page.
\begingroup
\setlength{\parskip}{0pt}
% Keep the short conclusion together when preceding content is compacted.
\Needspace{12\baselineskip}
\section{Conclusion}

We introduced TCSAlgBench, a benchmark and reusable pipeline for natural-language proof discovery in theoretical computer science. Its 398 challenges combine source context with expert-designed repairs for notation, interaction rules, and algorithm-design tasks. GPT-5.5, GPT-5.6 Sol, and Fable~5 demonstrate promising capabilities on these research-level tasks, with GPT models benefiting substantially from prover--verifier discussion. Our results identify prover--verifier discussion and repeated sampling as effective ways to improve coverage. In the workflow comparison using GPT-5.5, root-only decomposition provides further gains at a substantially lower token cost than agentic planning. When maximizing coverage takes priority over token cost, agentic planning achieves the highest five-run coverage among the tested workflows. Together, these findings highlight model capability and workflow design as complementary directions for improving automated TCS proof discovery.

%Evaluated systems access cited prior work, with source-paper proofs and lemmas withheld. Verifier-accepted coverage remains low despite refinement and repeated runs. Planning achieves the highest five-run coverage under matched call opportunities, but uses substantially more tokens than discussion. The pipeline provides a basis for evaluating successive models and agents on fresh, versioned challenges, with expert effort focused on construction and targeted validation.

\endgroup

%% file: sections/appendix_theorem-upper-bound.tex
% =========================================================================
% Research Challenge: Calibeating from online regret bounds
% Source paper: Calibeating Made Simple (arXiv:2603.22167), Theorem upper.bound
% License: CC BY 4.0 (https://creativecommons.org/licenses/by/4.0/)
%
% Self-contained appendix block: paste into your document body.
% Requires amsmath and amssymb; mathtools recommended (for \coloneqq).
% The \providecommand lines below supply the source paper's macros only
% if your preamble does not already define them.
% =========================================================================

\providecommand{\coloneqq}{\mathrel{:=}}
\providecommand{\bbR}{\mathbb{R}}
\providecommand{\E}{\mathbb{E}}
\providecommand{\simplex}{\Delta_K}
\providecommand{\calE}{\mathcal{E}}
\providecommand{\reg}{\mathrm{Reg}}

\section{Research Challenge: Calibeating from online regret bounds}\label{app:calibeating-challenge}

\noindent\textbf{Source paper:} \emph{Calibeating Made Simple},
\href{https://arxiv.org/abs/2603.22167}{arXiv:2603.22167}~\citep{chen2026calibeating} --- Theorem
\texttt{upper.bound}.\\
\textbf{License and adaptation:} The cited v1 source, by Yurong Chen,
Zhiyi Huang, Michael I. Jordan, and Haipeng Luo, is licensed under
\href{https://creativecommons.org/licenses/by/4.0/}{CC BY 4.0}; this
challenge excerpts and reformats definitions and notation and rephrases
Theorem \texttt{upper.bound} as an existence claim.\\
\textbf{Task:} Prove the theorem stated below. This document is
self-contained: all definitions and notation needed to understand the
statement are included. The proof is deliberately omitted.

\subsection*{Definitions and setup}

The following definitions and notation (quoted from the source paper) are
referenced by the theorem.

\paragraph{Model and notation.}
Let $K\geq 2$ be the number of possible outcomes, and
$\simplex \coloneqq \{p\in\bbR^K_{\geq 0}: \sum^K_{k=1} p_k = 1\}$ be the
probability simplex. The outcome space is denoted by
$\calE \coloneqq \{e_i: i \in [K]\} \subseteq \simplex$, where $e_i$ is the
$i$-th standard basis vector. We let $[n]$ denote the set $\{1,\dots,n\}$
for any positive integer $n$. Given a prediction sequence $p_{1:T}$ and
outcome sequence $y_{1:T}$, for any $p\in \simplex$, denote the number of
times the learner predicts $p$ as
$n_T(p) \coloneqq \sum^T_{t=1} \mathbf{1}\{p_t = p\}$, and the empirical
outcome distribution conditioned on prediction $p$ as
$\rho^p_{T}(y) \coloneqq \frac{1}{n_T(p)}\sum^T_{t=1}
\mathbf{1}\{p_t = p,\, y_t = y\}$ for $y\in\calE$, whenever $n_T(p)>0$.

\paragraph{Interaction protocol.}
The interaction proceeds for $T$ rounds. At each round $t \in [T]$, the
learner first observes $N$ external forecasts, $q^{(n)}_t\in \simplex$,
$n \in [N]$, and makes its own prediction $p_t \in \simplex$. The outcome
$y_t\in\calE$ is then revealed, and the learner incurs loss
$\ell(p_t,y_t)$. For simplicity, we assume that
$q_{1:T} \coloneqq (q_t)^T_{t=1}$ and $y_{1:T} \coloneqq (y_{t})^T_{t=1}$
are generated by an oblivious adversary, i.e., they are decided at time
$t=0$ with complete knowledge of the learner's algorithm (but not its
random bits).

\paragraph{Proper scoring loss.}
Throughout, we consider a proper scoring loss
$\ell: \Delta_K \times \calE \rightarrow \bbR$, i.e., losses such that for
any $q\in \Delta_K$,
$q\in \arg\min_{p \in \Delta_K} \E_{y\sim q}[\ell(p,y)]$. We write
$\ell(p,q)\coloneqq \E_{y\sim q}[\ell(p,y)]$.

\begin{description}
  \item[Definition (loss, refinement, calibration error).]
  Let $\ell$ be a proper scoring loss. The cumulative loss of predictions
  $p_{1:T}$ under outcomes $y_{1:T}$ is
  $L_T(p_{1:T},y_{1:T}) \coloneqq \sum^T_{t=1}\ell(p_t,y_t)$. The
  refinement score is
  $R_T(p_{1:T},y_{1:T}) \coloneqq \sum_{p} n_T(p)\,\ell(\rho^p_T, \rho^p_T)
  = \sum_{p}\min_{q \in \simplex}\sum_{t:p_t=p}\ell(q, y_t)$. Finally, the
  calibration error is
  $K_T(p_{1:T}, y_{1:T})\coloneqq L_T(p_{1:T},y_{1:T}) -
  R_T(p_{1:T},y_{1:T})$.

  \item[Definition (calibeating and multi-calibeating).]
  A learner is $\alpha(T)$-multi-calibeating w.r.t.\ loss $\ell$ if for any
  external forecasts $\{q^{(n)}_{1:T}\}^N_{n=1}$ and outcomes $y_{1:T}$,
  the learner's predictions $p_{1:T}$ satisfy
  $L_T(p_{1:T},y_{1:T}) \leq R_T(q^{(n)}_{1:T}, y_{1:T}) + \alpha(T)$ for
  all $n \in [N]$. We call $\alpha(T)$ the multi-calibeating rate. We say
  the learner is multi-calibeating if $\alpha(T)=o(T)$. When this holds in
  expectation over the learner's randomness, we call $\alpha(T)$ the
  expected multi-calibeating rate. When there is only $N=1$ external
  forecast, we simply say calibeating.

  \item[Definition (regret).]
  Define the regret of predictions $p_{1:T}$ under outcomes $y_{1:T}$ to be
  $\reg_T(p_{1:T},y_{1:T}) \coloneqq \sum_{t=1}^T \ell(p_t,y_t) -
  \min_{p\in\simplex}\sum_{t=1}^T \ell(p,y_t)$. We say an algorithm has
  (expected-)regret of $\alpha(T)$ if
  $\reg_T(p_{1:T},y_{1:T}) \leq \alpha(T)$ always holds (in expectation).

  \item[Definition ($Q$, distinct forecast values).]
  Let $Q\coloneqq \{q_t: t\in [T]\}$ denote the set of distinct external
  forecast values that appear over the horizon.
\end{description}

\paragraph{Context (companion lower bound).}
For any proper loss $\ell$, denote the optimal regret bound as
\begin{align*}
\beta(T) \coloneqq \inf_{\mathsf{A}} \sup_{y_{1:T}\in \calE^T}
\E_{p_{1:T}\sim \mathsf{A}}\left[\sum_{t=1}^T \ell(p_t, y_t)
- \min_{p \in \simplex} \sum_{t=1}^T \ell(p, y_t)\right],
\end{align*}
where $\mathsf{A}$ ranges over (possibly randomized) online algorithms.
Then every algorithm is at best
$|Q|\,\beta(\lfloor T/|Q|\rfloor)$-calibeating.

\subsection*{The challenge}

\noindent\textbf{Hypotheses / assumptions:}
\begin{itemize}
  \item $\ell$ is a proper loss;
  \item $\mathsf{A}$ is an online algorithm with regret $\alpha(T)$;
  \item $\alpha$ is a concave function.
\end{itemize}

\begin{quote}
\textbf{Statement to prove (Theorem \texttt{upper.bound}).}
\emph{For any proper loss $\ell$ and any online algorithm $\mathsf{A}$ with
regret $\alpha(T)$, where $\alpha$ is a concave function, there exists an
algorithm that is $|Q|\,\alpha(T/|Q|)$-calibeating.}
\end{quote}

\noindent\emph{(The original statement names the paper's specific
construction; it has been rephrased as an existence claim so that designing
the algorithm is part of the challenge.)}

\medskip
\noindent\emph{Free variables in the statement:} $\ell$, $\mathsf{A}$,
$\alpha$, $Q$, $T$.

\medskip
\noindent\textbf{Your task:} Prove the statement above, using only the
definitions and assumptions provided. State any additional standard
background results you invoke.

\paragraph{Note for readers (outside the example challenge).}
The material above is reproduced as an example challenge from our construction
pipeline. The companion lower bound provides context about the problem; it is
not needed to prove the target upper bound and does not reveal the algorithmic
construction or proof strategy required to establish it.

%% file: sections/appendix_dataset_construction_and_validation.tex
\section{Dataset Construction and Validation}\label{sec:construction}

This appendix details challenge construction, statement review, and versioning, followed by the supplementary pipeline for provisional Lean statements.

\subsection{Source Papers and Problem Selection}

The current candidate pool consists of papers accepted to STOC~2026 and COLT~2026 with public arXiv versions. We retrieve papers through the arXiv API and retain their TeX sources. Future candidate pools also require public arXiv sources and public releases additionally require CC BY 4.0 or CC0 licensing for the exact version used.

A model-assisted screen retains papers whose main results give upper or lower bounds on sample or runtime complexity and assigns each paper one primary area: algorithmic fairness and calibration, differential privacy, learning theory, optimization, sampling, or \emph{Other}. The last category includes graph and dynamic algorithms, quantum computing and information, coding theory, lattice-based cryptography, and fine-grained complexity or hardness of approximation. Within each paper, we select nontrivial theorem-level claims that can be made self-contained at moderate length.

\subsection{Proof-Graph Construction}

We use TeX sources to preserve theorem environments, labels, mathematical notation, and cross-references. A model pass identifies cited works available on arXiv, and a verifier checks for omissions before the references are resolved to TeX files. The source paper and resolved references are then converted into proof-dependency graphs.

Each numbered statement becomes a node containing its printed label, kind, verbatim statement, assumptions, free variables, and direct proof when present. A second pass records direct invocations of other result nodes as dependency edges. A glossary stores paper-specific terms, aliases, named objects, operators, functions, interaction rules, and algorithms with their verbatim definitions, excluding standard textbook background. Code attaches each glossary entry to nodes whose statements use its term or alias.

Deterministic checks enforce label consistency, absence of dangling references, and acyclicity; structural errors trigger regeneration. A semantic audit flags missed statements, mislabeled results, informal restatements, and suspicious dependencies for review. Untrusted cyclic or dangling edges are dropped and recorded. The source paper's graph is used only for challenge selection and is withheld from the prover; evaluation provides access to cited prior work through its graphs and TeX files.

\subsection{From Research Papers to Self-Contained Challenges}\label{sec:challenge-construction}

Each challenge contains a target theorem and the context needed to understand it, with the target proof withheld. Figure~\ref{fig:challenge-pipeline} gives an overview. The ten passes below are grouped into three phases; passes 2--10 apply expert-designed repair rules automatically in the stated order. Appendix~\ref{sec:prompt-challenge-refinement} provides representative prompts.

\begingroup
\setlength{\topsep}{2pt}
\setlength{\partopsep}{0pt}
\setlength{\parsep}{0pt}
\setlength{\itemsep}{2pt}
\paragraph{Select and assemble.}
\begin{enumerate}
    \item \textbf{Select statements and assemble the initial challenge.} An LLM selects one to four headline theorems and the definition and notation-setting IDs needed to interpret them. Code assembles their source text with a short LLM-generated overview. A self-containment checker returns catalogue IDs for missing context, preferring formal statements over informal versions; code inserts the selected text. This loop stops at a fixed round limit or when no new context can be added; gaps without a suitable entry remain recorded for later repair.
\end{enumerate}

\paragraph{Complete context.}
\begin{enumerate}
    \setcounter{enumi}{1}
    \item \textbf{Match glossary terms.} An LLM maps unresolved terms to source-glossary entries defining the same mathematical object, and code inserts the corresponding verbatim definitions using the returned indices. Unmatched terms remain unresolved.
    \item \textbf{Add standard definitions.} Add clearly labeled textbook background when a missing notion is genuinely standard.
    \item \textbf{Expand internal references.} Attach verbatim equations, assumptions, and algorithms referenced by the statement.
    \item \textbf{Resolve statement labels.} Replace raw \LaTeX{} cross-reference tokens with their resolved source labels.
    \item \textbf{Complete paper notation.} Insert available paper-specific notation definitions and normalize notation without changing its mathematical content.
    \item \textbf{Specify information access and action order.} Attach source descriptions of permitted observations, queries, actions, and their timing in online, oracle, streaming, or related settings.
\end{enumerate}

\paragraph{Finalize the challenge.}
\begin{enumerate}
    \setcounter{enumi}{7}
    \item \textbf{Harmonize assumption names.} Use consistent names for the same assumptions within a paper without changing their mathematical content.
    \item \textbf{Remove intermediate results.} Code strips source-paper lemma, proposition, claim, and corollary blocks from the draft.
    \item \textbf{Rewrite named constructions and check scope.} When algorithm design is part of the challenge, an LLM rewrites references to specific paper algorithms as existence claims and checks the intended scope of supplied black-box objects. Code also removes the corresponding algorithm pseudocode blocks so the final challenge withholds the construction.
\end{enumerate}
\endgroup

Mathematical context is quoted from the source, subject to notation normalization and the documented rewrites. Challenges may rely on standard TCS background; generated background definitions are explicitly labeled. Appendix~\ref{app:calibeating-challenge} reproduces a complete challenge from \emph{Calibeating Made Simple} produced by this pipeline.

\subsection{Statement Review and Versioning}

We developed the extraction, context-completion, repair, and self-containment rules using 21 challenges across topics. After the final repair pass, we manually reviewed 61 statements for fidelity and self-containment: the 21 development challenges and 40 additional challenges. All 40 additional challenges were judged faithful to the source results and self-contained. Headline model and agent comparisons use all 398 challenges, including the development challenges.

New releases add versioned batches while preserving prior evaluation sets. Future batches can draw on newly released papers, including FOCS and SODA papers, using the same construction and validation pipeline. Each release records new and unchanged challenges and their evaluation splits.

We record each challenge's first arXiv version date and partition results by assumed model-family knowledge cutoffs. This is a coarse diagnostic: publication dates alone cannot rule out contamination. Appendix~\ref{app:model-results} reports the source-date results.

\subsection{Provisional Lean Statement Construction}

\paragraph{Generation.}
A GPT-5.5 xhigh formalization agent receives the self-contained challenge and a Mathlib/CSLib API guide and writes a \texttt{def name\_statement : Prop} declaration, adding minimal local definitions where needed. Declarations must compile in the pinned Lean environment without \texttt{sorry}, \texttt{admit}, or untrusted axioms.

\paragraph{Semantic checks.}
Compilation alone cannot establish fidelity. A separate GPT-5.5 high-effort call back-translates the Lean statement into natural language without seeing the source statement. The consistency gate (C) compares the challenge with this back-translation, checking hypotheses, quantifiers, conclusions, bounds, and boundary cases; its jurors do not see the Lean code. The degeneracy gate (D) receives the challenge and Lean code and checks for vacuous assumptions, trivial witnesses, degenerate Mathlib values, and other ways to satisfy the encoding without the intended mathematical content.

\paragraph{Voting.}
Each gate uses three GPT-5.5 high-effort and three Opus~4.8 high-effort jurors. Acceptance requires at least two yes votes within each model family on both gates. Jurors are stateless and see neither earlier reviews nor one another's judgments; parsing failures and call errors count as no votes. If exactly one family fails a gate by one vote, its three-juror panel is resampled once for that gate and turn, retaining whichever panel has more yes votes. The acceptance rule is then reapplied.

\paragraph{Revision and retention.}
The agent receives up to five rounds of generation and review, stopping on acceptance. After failed reviews, dissent guides revision of the best candidate so far, ranked first by gates passed and then by total yes votes. The pipeline retains 221 compiler-valid candidates from 398 challenges (55.5\%); Table~\ref{tab:lean-statements-by-topic} reports retention by topic.

\begin{table}[t]
    \centering
    \small
    \caption{Candidate Lean-statement retention by topic after compilation and automated semantic filtering. The rows follow the topic order used in Table~\ref{tab:all-topic}. Retention is a construction diagnostic, not expert-validated formalization accuracy.}
    \label{tab:lean-statements-by-topic}
    \begin{tabular}{lrr}
        \hline
        Topic & Retained & Total \\
        \hline
        Algorithmic fairness & 17 & 30  \\
        Differential privacy & 11 & 33  \\
        Learning theory      & 61 & 101 \\
        Optimization         & 25 & 52  \\
        Sampling             & 22 & 38  \\
        Other                & 85 & 144 \\
        \hline
        Total                & 221 & 398 \\
        \hline
    \end{tabular}
\end{table}

%% file: sections/appendix_evaluation_protocol.tex
\section{Evaluation Protocol Details}\label{sec:evaluation_protocol}

This appendix describes external verification, repeated-run metrics, and diagnostic procedures. All headline model and agent acceptance comparisons use the full collection of 398 challenges.

\subsection{External Verification and Author Inspection}

Each system submits one final natural-language proof. Before external verification, a proof-organizing step checks whether statements cited from prior work match their original source statements. Three separately sampled GPT-5.5 high-effort voters receive only the problem statement and rewritten proof and return PASS or FAIL with a rationale. At least two PASS votes are required for acceptance. A system's internal verifier is part of the workflow being evaluated and never contributes to the benchmark label.

An author reviewed 10 accepted proofs produced by GPT-5.5 xhigh through prover--verifier discussion on challenges in the 61-problem manually reviewed statement set. All 10 accepted proofs were correct. We also re-score identical submissions with an alternative Opus~4.8 high-effort verifier. Section~\ref{sec:robustness-results} reports the resulting coverage counts and changes in ordering.

\paragraph{Empirical precedents for automated verification.}\label{sec:verification-precedents}
BrokenMath reports 95\% agreement with human annotations on 250 responses to false mathematical statements~\citep{petrovDV25}. QED combines separate prover and verifier calls with staged structural and detailed checks, producing expert-verified proofs for three open research problems~\citep{anYPZ26}. Following prompt optimization, concurrent TCS-BENCH reports over 90\% accuracy against expert judgments on 100 human-labeled proofs~\citep{cohenAddadEtAl26tcsbench}. Reported agreement depends on the task and judging protocol, including TCS-BENCH's access to a reference proof; it does not directly validate our statement-and-proof-only verifier.

\subsection{Agent Workflow Definitions}\label{app:agent-workflows}

We compare four workflows using GPT-5.5 xhigh, drawing on recurring proof-agent mechanisms~\citep{thakurEtAl24copra,zhangEtAl25cumulative,chungEtAl26goedelarchitect,liEtAl26agentflow,liuEtAl26danus,zhangEtAl26valg,tsoukalasEtAl26alphaproofnexus}. Figure~\ref{fig:agent-workflows} summarizes their information flow.

\paragraph{Discussion.}
A decider, prover, and verifier share a bounded discussion history. The prover revises its argument using verifier feedback, without explicit decomposition~\citep{fengEtAl26,anYPZ26,schmittEtAl26proofcouncil,zhaoYCC26,huangYang25imopipeline,balkoEtAl26bolzano}.

\paragraph{Root-only decomposition.}
Each attempt creates one depth-1 dependency-aware lemma plan, proves its subgoals, and recombines them. After failure, the workflow requests a new root-level plan~\citep{jiangEtAl23dsp,corneli26firstproofsprint,varamballyEtAl25hilbert}.

\paragraph{Decomposition with MCTS.}
The search maintains an AND--OR tree of proof goals~\citep{kungSHYLSOLLGLPP26,wangEtAl24poetry}. An OR-node represents a goal that can be established by an accepted direct proof or a successful decomposition. An AND-node represents a decomposition whose child claims must all be established. Nodes track their search status, estimated value, importance, visit count, and attempt history. At a selected open goal, a decider coordinates proof, disproof, or expansion attempts through the prover--verifier loop. Proposed reductions undergo checks for variable scope, assumptions, and non-triviality, followed by verification of their recombination arguments before attachment to the tree.

Selection uses estimated values~\citep{lampleEtAl22htps,xinEtAl25deepseek,hubertEtAl26alphaproof}. We collect open goals along promising decomposition paths and, when they exceed the per-iteration limit, rank them by importance-weighted upper confidence bounds:
\[
    \operatorname{score}(v)
    =
    i(v)\left[
   \widehat q(v)
    +c\sqrt{\frac{\log(1+n_{\mathrm{ref}})}{1+n(v)}}
    \right],
\]
where $n(v)$ is the visit count of goal $v$, $n_{\mathrm{ref}}$ is the total visit count across candidate goals, $c$ controls exploration, and $i(v)$ represents the goal's share of the root's proof difficulty. The root has importance one. For each new decomposition, an LLM assigns normalized difficulty shares to its children, which inherit their parent's importance multiplicatively. Failed importance calls use uniform shares.

The evaluator assesses completeness $\kappa(v)$, the importance-weighted fraction of difficulty already closed by proved or dead parts, and solvability, the estimated likelihood of closing the remaining work. Its search value is
\[
    \widehat q(v)
    =
    s(v)\,
    \frac{1+\kappa(v)}{2},
\]
where $s(v)$ is the solvability estimate.

After each iteration, outcomes and refreshed values propagate through the ancestors, and visit counts increase. Proved subgoals redistribute their importance among still-open siblings. If any child becomes dead, its decomposition becomes dead and that decomposition's subgoal importances become zero. Search continues until the root closes or the budget in Appendix~\ref{app:budgets-reporting} is exhausted. The assembled proof receives separate external verification.

\paragraph{Agentic planning.}
A planner constructs and revises a lemma DAG while coordinating decomposition, subgoal proving, critique, and recombination. After a failed attempt, it receives failure feedback and the current proof state to revise the DAG~\citep{anYPZ26,wuEtAl26starpolya,kripner26openprover,xinEtAl25bfsproverv2}.

\paragraph{Search diagnostics.}
For decomposition-based workflows, only the final recombined proof contributes to acceptance. We record reduction validity, whether subgoals are simpler than the root theorem, which subgoals are accepted, and recombination soundness to distinguish failures of planning, subproblem solving, and assembly.

\subsection{Budgets and Reporting}\label{app:budgets-reporting}

All runs use the fixed offline corpus and access restrictions in Section~\ref{sec:main-evaluation-protocol}, with a 128K-token per-call output cap. The public release represents cited documents through a versioned manifest.

The model comparison pairs single-call direct inference with 10-round prover--verifier discussion. The separate agent comparison fixes GPT-5.5 xhigh, problem input, tool access, and the external verifier, and matches model-call opportunities across workflows. Each challenge--seed run permits at most 20 outer iterations, with up to six proof goals attempted in parallel and at most 10 discussion rounds per goal within each iteration. Calls by every role count toward the budget.

Seed-1 acceptance is the fraction of the 398 challenges accepted in the first run. Multi-run coverage counts a challenge once if any independently seeded run is accepted. Direct inference uses ten runs; model-comparison discussion and the agent workflows each use five runs.

Input and output tokens are reported separately as realized usage because workflows retain or revisit different amounts of context. For each agent workflow, token counts are averaged per challenge--seed run over 10 randomly selected challenges and five independent seeds.

\subsection{Confidence Calibration}

Calibration uses a separate run for each model on all 398 challenges. Each model reports confidence $p_i\in[0,1]$ during proof generation, before external evaluation; we pair it with that proof's acceptance label. Following Humanity's Last Exam~\citep{phan2025humanitys}, we compute root mean square calibration error from confidence bins $B_1,\ldots,B_K$:
\[
    \operatorname{RMSCE}
    =
    \sqrt{\sum_{k=1}^{K}\frac{|B_k|}{N}
    \left(\operatorname{acc}(B_k)-\operatorname{conf}(B_k)\right)^2}.
\]
Here $N=398$, $\operatorname{acc}(B_k)$ is the bin's verifier-acceptance rate, and $\operatorname{conf}(B_k)$ is its mean reported confidence. We use the released implementation with $p=2$ and $\beta=40$. Table~\ref{tab:calibration-results} reports acceptance and calibration from these same runs; calibration targets individual-proof acceptance, not union-of-runs coverage.

% \subsection{Lean Proof Diagnostic}

% The formalization diagnostic fixes a retained candidate statement and supplies its natural-language reference proof. We test direct proof generation and iterative compiler-guided repair in a pinned Lean environment. A submitted file must compile and may not use \texttt{sorry}, \texttt{admit}, untrusted axioms, or modifications to the candidate statement. We record parse and elaboration errors, unsolved goals, timeouts, forbidden constructs, and statement modification. Kernel acceptance would establish only the provisional candidate declaration; it would not certify that the declaration faithfully represents the source theorem.

%% file: sections/appendix_secondary_results.tex
\section{Detailed Results and Diagnostics}\label{app:secondary}

This appendix reports the numerical results behind Section~\ref{sec:evaluation}. Model and agent acceptance comparisons use all 398 challenges; the Lean diagnostic uses the 221 retained candidate statements.

\input{sections/appendix_benchmark_comparison}

\subsection{Complete Model Results}\label{app:model-results}

\paragraph{Direct inference.}
Fable~5 max is excluded due to output-token access limits.

\begin{table}[H]
    \centering
    \small
    \caption{Direct-inference results for Figure~\ref{fig:model-comparison}(a) on all 398 challenges, using three-voter majority voting. Entries are accepted-challenge counts, with percentages in parentheses. Seed~1 reports one run; 10-run coverage counts challenges accepted in any of ten independently seeded runs. Input and output tokens are averaged per prover call; K denotes thousands.}
    \label{tab:direct-inference-baseline}
    \begin{tabular}{lcccc}
        \hline
        Configuration & Seed~1 acceptance & 10-run coverage & Avg. input tokens & Avg. output tokens \\
        \hline
        GPT-5.6 Sol max   & 23 (5.8\%) & 55 (13.8\%) & 17K  & 19K \\
        GPT-5.6 Sol xhigh & 19 (4.8\%) & 43 (10.8\%) & 17K  & 16K \\
        GPT-5.6 Sol high  & 19 (4.8\%) & 41 (10.3\%) & 16K  & 10K \\
        GPT-5.5 xhigh     & 13 (3.3\%) & 28 (7.0\%)  & 15K  & 17K \\
        GPT-5.5 high      & 8 (2.0\%)  & 21 (5.3\%)  & 16K  & 16K \\
        Opus~4.8 max      & 3 (0.8\%)  & 7 (1.8\%)   & 20K  & 9K \\
        Opus~4.8 xhigh    & 1 (0.3\%)  & 6 (1.5\%)   & 20K  & 8K \\
        Opus~4.8 high     & 1 (0.3\%)  & 4 (1.0\%)   & 21K  & 7K \\
%        Fable~5 max       & 16 (4.0\%) & 32 (8.0\%)  & 28K  & 90K \\
        Fable~5 xhigh     & 27 (6.8\%) & 45 (11.3\%) & 27K  & 62K \\
        Fable~5 high      & 18 (4.5\%) & 58 (14.6\%) & 25K  & 46K \\
        \hline
    \end{tabular}
\end{table}

\paragraph{Multi-round results by topic.}
The topic breakdown in Table~\ref{tab:all-topic} underlies Figure~\ref{fig:model-comparison}(b), with GPT-5.6 Sol max's five-run results plotted in Figure~\ref{fig:topic-coverage}. Across the eight non-Fable configurations, one full 10-round discussion run averages approximately 453.7K input tokens and 51.4K output tokens per challenge.

\begin{table}[H]
    \centering
    \caption{Verifier-accepted proofs by topic on all 398 TCSAlgBench challenges after 10-round discussion. Panel A reports seed~1, and Panel B reports five-run coverage. AF denotes algorithmic fairness; DP, LT, Opt., and Samp. abbreviate differential privacy, learning theory, optimization, and sampling, respectively. Parenthesized column-header values give category totals.}
    \label{tab:all-topic}
    \resizebox{\textwidth}{!}{%
    \begin{tabular}{lrrrrrrr}
        \hline
        Configuration & AF (30) & DP (33) & LT (101) & Opt. (52) & Samp. (38) & Other (144) & All (398) \\
        \hline
        \multicolumn{8}{l}{\emph{Panel A: First run (seed 1; $N=398$)}} \\
        Opus~4.8 high     & 3  & 0 & 2  & 1  & 1  & 1  & 8 \\
        Opus~4.8 xhigh    & 4  & 0 & 2  & 1  & 2  & 1  & 10 \\
        Opus~4.8 max      & 3  & 0 & 2  & 1  & 2  & 1  & 9 \\
        GPT-5.5 high      & 9  & 1 & 9  & 3  & 6  & 14 & 42 \\
        GPT-5.5 xhigh     & 8  & 1 & 12 & 5  & 6  & 13 & 45 \\
        GPT-5.6 Sol high  & 11 & 1 & 13 & 8  & 7  & 14 & 54 \\
        GPT-5.6 Sol xhigh & 10 & 1 & 15 & 9  & 8  & 21 & 64 \\
        GPT-5.6 Sol max   & 11 & 1 & 17 & 10 & 7  & 29 & 75 \\
        Fable~5 high      & 8  & 0 & 7  & 7  & 6  & 9  & 37 \\
        Fable~5 xhigh     & 9  & 1 & 8  & 6  & 3  & 11 & 38 \\
        \hline
        \multicolumn{8}{l}{\emph{Panel B: Five-run coverage ($N=398$)}} \\
        Opus~4.8 high     & 4  & 0 & 2  & 2  & 3  & 3  & 14 \\
        Opus~4.8 xhigh    & 6  & 0 & 2  & 2  & 3  & 2  & 15 \\
        Opus~4.8 max      & 5  & 1 & 3  & 2  & 4  & 3  & 18 \\
        GPT-5.5 high      & 10 & 1 & 17 & 8  & 8  & 22 & 66 \\
        GPT-5.5 xhigh     & 12 & 1 & 21 & 8  & 7  & 23 & 72 \\
        GPT-5.6 Sol high  & 13 & 1 & 18 & 11 & 10 & 26 & 79 \\
        GPT-5.6 Sol xhigh & 14 & 1 & 22 & 12 & 10 & 29 & 88 \\
        GPT-5.6 Sol max   & 12 & 2 & 25 & 12 & 8  & 35 & 94 \\
        Fable~5 high      & 9  & 1 & 14 & 9  & 9  & 20 & 62 \\
        Fable~5 xhigh     & 10 & 1 & 15 & 10 & 9  & 16 & 61 \\
        \hline
    \end{tabular}%
    }
\end{table}

\paragraph{Alternative-verifier results.}
Table~\ref{tab:verifier-sensitivity} gives the model-level counts summarized in Section~\ref{sec:robustness-results}.

\begin{table}[H]
    \centering
    \small
    \caption{Five-run coverage for the eight model configurations evaluated under both the primary three-voter majority rule and an alternative Opus~4.8 verifier, using the same generated proofs. $\Delta$ is Opus-verifier coverage minus majority-vote coverage.}
    \label{tab:verifier-sensitivity}
    \begin{tabular}{lrrr}
        \hline
        Configuration & Three-voter majority & Opus~4.8 verifier & $\Delta$ \\
        \hline
        Opus~4.8 high     & 14 & 14 & 0 \\
        Opus~4.8 xhigh    & 15 & 16 & +1 \\
        Opus~4.8 max      & 18 & 18 & 0 \\
        GPT-5.5 high      & 66 & 60 & -6 \\
        GPT-5.5 xhigh     & 72 & 68 & -4 \\
        GPT-5.6 Sol high  & 79 & 77 & -2 \\
        GPT-5.6 Sol xhigh & 88 & 90 & +2 \\
        GPT-5.6 Sol max   & 94 & 93 & -1 \\
        \hline
    \end{tabular}
\end{table}

\paragraph{Results around model knowledge cutoffs.}
We partition challenges by their first arXiv date relative to each model family's assumed cutoff (Table~\ref{tab:scale-cutoff}); Figure~\ref{fig:cutoff-analysis} plots the five-run results.

\begin{table}[H]
    \centering
    \caption{Verifier-accepted proofs before and after the assumed model-family knowledge cutoff on all 398 challenges. Panel A reports seed~1, and Panel B reports five-run coverage. Percentages use the corresponding pre- or post-cutoff column total; percentages in the All column use all 398 challenges.}
    \label{tab:scale-cutoff}
    \small
    \renewcommand{\arraystretch}{1.10}
    \begin{tabular*}{\textwidth}{@{\extracolsep{\fill}}lclll@{}}
        \hline
        Models & Cutoff & Pre-cutoff & Post-cutoff & All \\
        \hline
        \multicolumn{5}{l}{\emph{Panel A: First run (seed 1; $N=398$)}} \\
        Opus~4.8 high  & 2026-01-01 & 3/201 (1.5\%)   & 5/197 (2.5\%)   & 8/398 (2.0\%) \\
        Opus~4.8 xhigh & 2026-01-01 & 3/201 (1.5\%)   & 7/197 (3.6\%)   & 10/398 (2.5\%) \\
        Opus~4.8 max   & 2026-01-01 & 2/201 (1.0\%)   & 7/197 (3.6\%)   & 9/398 (2.3\%) \\
        GPT-5.5 high   & 2025-12-01 & 19/188 (10.1\%) & 23/210 (11.0\%) & 42/398 (10.6\%) \\
        GPT-5.5 xhigh  & 2025-12-01 & 24/188 (12.8\%) & 21/210 (10.0\%) & 45/398 (11.3\%) \\
        GPT-5.6 Sol high  & 2026-02-16 & 33/262 (12.6\%) & 21/136 (15.4\%) & 54/398 (13.6\%) \\
        GPT-5.6 Sol xhigh & 2026-02-16 & 41/262 (15.6\%) & 23/136 (16.9\%) & 64/398 (16.1\%) \\
        GPT-5.6 Sol max   & 2026-02-16 & 46/262 (17.6\%) & 29/136 (21.3\%) & 75/398 (18.8\%) \\
        Fable~5 high      & 2026-01-01 & 17/201 (8.5\%)  & 20/197 (10.2\%) & 37/398 (9.3\%) \\
        Fable~5 xhigh     & 2026-01-01 & 18/201 (9.0\%)  & 20/197 (10.2\%) & 38/398 (9.5\%) \\
        \hline
        \multicolumn{5}{l}{\emph{Panel B: Five-run coverage ($N=398$)}} \\
        Opus~4.8 high  & 2026-01-01 & 6/201 (3.0\%)   & 8/197 (4.1\%)   & 14/398 (3.5\%) \\
        Opus~4.8 xhigh & 2026-01-01 & 6/201 (3.0\%)   & 9/197 (4.6\%)   & 15/398 (3.8\%) \\
        Opus~4.8 max   & 2026-01-01 & 7/201 (3.5\%)   & 11/197 (5.6\%)  & 18/398 (4.5\%) \\
        GPT-5.5 high   & 2025-12-01 & 31/188 (16.5\%) & 35/210 (16.7\%) & 66/398 (16.6\%) \\
        GPT-5.5 xhigh  & 2025-12-01 & 38/188 (20.2\%) & 34/210 (16.2\%) & 72/398 (18.1\%) \\
        GPT-5.6 Sol high  & 2026-02-16 & 53/262 (20.2\%) & 26/136 (19.1\%) & 79/398 (19.8\%) \\
        GPT-5.6 Sol xhigh & 2026-02-16 & 59/262 (22.5\%) & 29/136 (21.3\%) & 88/398 (22.1\%) \\
        GPT-5.6 Sol max   & 2026-02-16 & 58/262 (22.1\%) & 36/136 (26.5\%) & 94/398 (23.6\%) \\
        Fable~5 high      & 2026-01-01 & 29/201 (14.4\%) & 33/197 (16.8\%) & 62/398 (15.6\%) \\
        Fable~5 xhigh     & 2026-01-01 & 28/201 (13.9\%) & 33/197 (16.8\%) & 61/398 (15.3\%) \\
        \hline
    \end{tabular*}
\end{table}

This aggregate split cannot establish training-set membership or rule out contamination; the cohorts may also differ in topic and difficulty.

\paragraph{Complementarity across effort settings.}
Table~\ref{tab:effort-complementarity} shows that effort settings reach non-nested challenge sets.

\begin{table}[H]
    \centering
    \small
    \setlength{\tabcolsep}{4pt}
    \renewcommand{\arraystretch}{1.10}
    \caption{Complementarity of five-run verifier-accepted coverage on all 398 challenges, partitioned by the model-family knowledge cutoff. Overlap counts challenges accepted under both effort settings, while the setting-only columns count challenges accepted uniquely under one setting. For Fable~5, the high/xhigh union contains 32 challenges before the cutoff, 37 after it, and 69 overall.}
    \label{tab:effort-complementarity}
    \begin{tabular*}{\textwidth}{@{\extracolsep{\fill}}lrrrrrr@{}}
        \hline
        \multicolumn{7}{l}{\emph{Panel A: Opus~4.8 and GPT-5.5}} \\
        & \multicolumn{3}{c}{Opus~4.8: xhigh vs.\ max}
        & \multicolumn{3}{c}{GPT-5.5: high vs.\ xhigh} \\
        \cline{2-4}\cline{5-7}
        & {\scriptsize Overlap} & {\scriptsize xhigh only} & {\scriptsize max only}
        & {\scriptsize Overlap} & {\scriptsize high only} & {\scriptsize xhigh only} \\
        \hline
        Before cutoff & 5  & 1 & 2 & 29 & 2 & 9  \\
        After cutoff  & 9  & 0 & 2 & 31 & 4 & 3  \\
        Total         & 14 & 1 & 4 & 60 & 6 & 12 \\
        \hline
        \multicolumn{7}{l}{\emph{Panel B: GPT-5.6 Sol and Fable~5}} \\
        & \multicolumn{3}{c}{GPT-5.6 Sol: xhigh vs.\ max}
        & \multicolumn{3}{c}{Fable~5: high vs.\ xhigh} \\
        \cline{2-4}\cline{5-7}
        & {\scriptsize Overlap} & {\scriptsize xhigh only} & {\scriptsize max only}
        & {\scriptsize Overlap} & {\scriptsize high only} & {\scriptsize xhigh only} \\
        \hline
        Before cutoff & 51 & 8  & 7  & 25 & 4 & 3 \\
        After cutoff  & 27 & 2  & 9  & 29 & 4 & 4 \\
        Total         & 78 & 10 & 16 & 54 & 8 & 7 \\
        \hline
    \end{tabular*}
\end{table}

\subsection{Complete Agent-Design Results}\label{app:agent-results}

\begin{table}[H]
    \centering
    \small
    \setlength{\tabcolsep}{5pt}
    \renewcommand{\arraystretch}{1.12}
    \caption{Call-matched agent-design comparison using GPT-5.5 xhigh and the same external verifier; Appendix~\ref{sec:evaluation_protocol} details the budgets. Acceptance counts use all 398 challenges, with rates in parentheses. Seed~1 reports one run; five-run coverage counts challenges accepted in any of five independently seeded runs. Tokens are averaged per challenge--seed run over 10 randomly selected challenges and five independent seeds; M denotes millions.}
    \label{tab:preliminary-agent-ablation}
    \resizebox{\textwidth}{!}{%
    \begin{tabular}{@{}lrrrr@{}}
        \hline
        Workflow & Seed~1 & Five-run coverage & Avg. input tokens & Avg. output tokens \\
        \hline
        Decomposition with MCTS search
        & 77 (19.3\%) & 96 (24.1\%) & 27.4M & 3.9M \\
        Discussion, no decomposition
        & 60 (15.1\%) & 84 (21.1\%) & 12.1M & 1.0M \\
        Discussion, root-only decomposition
        & 73 (18.3\%) & 93 (23.4\%) & 25.6M & 3.9M \\
        Discussion, agentic planning
        & 72 (18.1\%) & 101 (25.4\%) & 60.4M & 9.0M \\
        \hline
    \end{tabular}
    }
\end{table}

Figure~\ref{fig:agent-comparison} plots these results. In the token-usage sample, agentic planning processes approximately five times the input and nine times the output tokens of discussion without decomposition.

\subsection{Confidence and Verifier Acceptance}\label{app:calibration-results}

We measure confidence calibration to external verifier acceptance using the procedure in Appendix~\ref{sec:evaluation_protocol}.

\begin{table}[H]
    \centering
    \caption{Verifier acceptance and RMS calibration error (RMS-CE), in percent, on all 398 challenges. These runs are separate from the main model comparison. Both metrics use the same proof outputs, with confidence elicited during proof generation before external verification.}
    \label{tab:calibration-results}
    \begin{tabular}{lrr}
        \hline
        Model configuration & Acceptance (\%) $\uparrow$ & RMS-CE (\%) $\downarrow$ \\
        \hline
        GPT-5.6 Sol max       & 18.1 & 10.8 \\
        GPT-5.6 Sol xhigh     & 15.6 & 4.3 \\
        GPT-5.6 Sol high      & 12.1 & 11.7 \\
        GPT-5.5 xhigh         & 10.8 & 19.1 \\
        GPT-5.5 high          & 10.1 & 17.3 \\
        Claude Opus~4.8 xhigh & 2.3  & 2.0 \\
        Claude Opus~4.8 high  & 2.0  & 1.7 \\
        Claude Opus~4.8 max   & 1.8  & 2.5 \\
        \hline
    \end{tabular}
\end{table}

Acceptance and calibration rankings differ. The low Opus~4.8 errors are consistent with assigning confidence near a low acceptance base rate; they do not show that the model can identify its rare accepted proofs. Reliability diagrams, Brier scores, and expert labels are needed for a stronger proof-level interpretation.

\subsection{Prospective Formalization Resource}\label{app:formalization-results}

The construction pipeline retains 221 compiler-valid Lean declarations from 398 natural-language challenges (55.5\%) after the automated semantic checks in Section~\ref{sec:lean-formalization}. Table~\ref{tab:lean-statements-by-topic} gives the topic breakdown. The tested baseline agents produce zero accepted sorry-free Lean proofs on these candidates, which remain a prospective resource for formalization research.

\emph{Beyond the Library}~\citep{moakhar2026beyond} reports expert-validated statement-and-proof formalizations of seven selected papers, including five STOC papers, with public artifacts; two developments require no axioms beyond Lean's kernel. These case studies demonstrate feasibility in selected settings and motivate expert audit of our candidates. Differences in task selection, context, libraries, axiom policies, and human validation make the results complementary rather than directly comparable.

%% file: sections/appendix_benchmark_comparison.tex
\subsection{Context from Published Mathematics Benchmarks}\label{app:benchmark-comparison}

Table~\ref{tab:benchmark-context} records MathArena's published results as of September 18, 2026, for the two configurations also evaluated here. These benchmarks report average final-answer accuracy over repeated attempts; ArXivMath likewise checks research-derived answers rather than proof correctness. TCSAlgBench reports five-run proof coverage under 10-round prover--verifier discussion (Table~\ref{tab:all-topic}, Panel B), so the metrics and evaluation protocols differ.

\begin{table}[H]
    \centering
    \small
    \caption{Published MathArena results and TCSAlgBench five-run coverage with 10-round discussion, in percent. Metrics and evaluation protocols differ.}
    \label{tab:benchmark-context}
    \resizebox{\textwidth}{!}{%
    \begin{tabular}{llrr}
        \hline
        Benchmark & Metric & GPT-5.5 xhigh & Opus~4.8 max \\
        \hline
        AIME 2026 & Final-answer accuracy & 100.00 & 100.00 \\
        HMMT February 2026 & Final-answer accuracy & 98.48 & 95.45 \\
        Apex & Final-answer accuracy & 80.21 & 81.25 \\
        Apex Shortlist & Final-answer accuracy & 98.40 & 90.43 \\
        ArXivMath June 2026 & Final-answer accuracy & 83.63 & 69.97 \\
        \hline
        TCSAlgBench & Five-run proof coverage & 18.1 & 4.5 \\
        \hline
    \end{tabular}%
    }
\end{table}

%% file: sections/appendix_prompts.tex
\section{Prompt Templates}\label{sec:prompt-templates}

This appendix records representative construction and proof-agent prompts. Instruction wording is preserved, and the external verifier's framing adjustment is noted in Appendix~\ref{sec:prompt-verifier}. Double-braced names mark runtime substitutions. \texttt{[SYSTEM]}, \texttt{[USER]}, and \texttt{[OMITTED: ...]} are editorial labels; omissions are identified where used.

\subsection{Challenge Generation Pipeline}

\subsubsection{Challenge Selection and Assembly}\label{sec:prompt-challenge-selection}

The selector receives the extracted statement catalogue and returns headline theorem IDs and the context IDs needed to understand them. Code assembles the corresponding source text after proof-graph extraction (Appendix~\ref{sec:construction}).

\Needspace{5\baselineskip}
\begin{lstlisting}[style=benchmarkprompt]
[SYSTEM]

You are a mathematician curating a benchmark of self-contained research challenges from a paper's formal statements. You return ONE JSON object -- no prose around it.

You are given the full list of the paper's numbered statements (theorems, lemmas, propositions, corollaries, definitions), each with an id, kind, and verbatim text.

Your job:
  1. Select the paper's MAIN / headline theorems -- the central results that represent the paper's contribution. Usually 1 to 4. Prefer items of kind "theorem" (occasionally a headline "proposition" or "corollary" if that is the paper's main result). Do NOT select minor intermediate lemmas, technical corollaries, or helper results.
  2. For each selected theorem, list the ids of the DEFINITIONS and notation-setting statements (from the provided list) that a reader needs in order to UNDERSTAND THE THEOREM STATEMENT itself -- the objects, quantities, and terms it references. Include only what is needed to make the statement self-contained; keep it minimal but complete. These are typically kind "definition" (and occasionally a setup lemma/proposition that defines an object the theorem uses). Do NOT include results used only in the proof.
  3. Write a 1-2 sentence plain-language "intuition" describing what the theorem claims, in your own words. Do not include any proof.

Output schema:
{
  "main_theorems": [
    {
      "theorem_id": "<id from the list>",
      "title": "<short human-readable title, <= 12 words>",
      "definition_ids": ["<id>", "<id>", ...],
      "intuition": "<1-2 sentence plain-language description of the claim>"
    },
    ...
  ]
}
If the paper has no clear main theorem (e.g. only definitions were extracted), return
{"main_theorems": []}.

[USER]

## Paper
title: {{PAPER_TITLE}}
arxiv: {{ARXIV_ID}}

## Statements
{{STATEMENT_CATALOGUE}}

## Your task
Return the JSON object selecting the paper's main theorems, the definition ids each needs to be self-contained, and a short intuition for each. Use ONLY ids from the list above.
\end{lstlisting}

\subsubsection{Challenge Refinement}\label{sec:prompt-challenge-refinement}

The templates below cover self-containment, glossary matching, and existence rewriting within the expert-designed sequence in Appendix~\ref{sec:challenge-construction}. Other passes complete references, notation, information access and action order, and assumption names. Intermediate-result removal is deterministic and has no LLM prompt.

\paragraph{Iterative self-containment check.}
The checker receives the target theorem, included definitions, and remaining statement catalogue. Its \texttt{missing} entries guide context additions; gaps without an available definition remain recorded for later repair.

\Needspace{5\baselineskip}
\begin{lstlisting}[style=benchmarkprompt]
[SYSTEM]

You are a meticulous mathematical referee checking whether a 'research challenge' document is SELF-CONTAINED. You return ONE JSON object -- no prose around it.

The document presents a theorem to prove, preceded by a set of included definitions. A reader should be able to understand the THEOREM STATEMENT (and the included definitions) using only (a) what is included in the document and (b) standard general mathematical background.

You are given:
  - the target theorem (id + verbatim text + assumptions),
  - the definitions currently INCLUDED (ids + verbatim text),
  - a CATALOGUE of all OTHER statements available from the same paper (ids + kind + text), which may be added if needed.

Your job: find every term, object, operator, quantity, or piece of notation that the theorem statement (or an included definition) RELIES ON to be understood, that is NOT defined in the included material and is NOT standard general background.

What counts as STANDARD GENERAL BACKGROUND (do NOT flag these): real/integer/natural numbers, sets, functions, sup/inf/min/max, expectation/probability/variance, asymptotic notation (O, Omega, Theta, O-tilde, o), norms (ell_p, Euclidean), inner products, convexity/concavity, Lipschitzness, gradients/Hessians, VC dimension, PAC learning basics, standard distributions (Gaussian, uniform), KL divergence, entropy, big-name standard objects a graduate reader knows.

What to FLAG (paper-specific, must be defined in the doc): bespoke quantities and scores the paper defines (e.g. a custom "refinement score", "calibeating rate", "rho-replicable"), named algorithms the statement refers to (e.g. "Algorithm 1"), paper-specific operators or function classes, non-standard notation introduced earlier in the paper.

For each flagged gap, if the CATALOGUE contains a statement that defines it, give that id in "add_id". If no available statement defines it (it was defined inline in prose the extraction didn't capture, or it refers to an algorithm/figure), set "add_id" to "" and briefly say so.

FORMAL OVER INFORMAL: papers often state a definition/theorem twice -- a loose INFORMAL version (flagged "informal", "(... ; informal, see Definition/Theorem N)", "Informal Definition/Theorem", or in an intro/overview) and a precise FORMAL version. When BOTH are in the catalogue, always choose the FORMAL one's id in "add_id". If the document currently INCLUDES the informal version, flag it: set its term, note "included version is informal", and put the formal version's catalogue id in "add_id" so it gets swapped in.

Output schema:
{
  "complete": <true|false>,            // true if the doc is self-contained (no real gaps)
  "missing": [
    {"term": "<the undefined term/notation>",
     "why_needed": "<where/how the theorem relies on it>",
     "add_id": "<catalogue id that defines it, or '' if none available>"}
  ],
  "notes": "<optional: anything the author should know>"
}
Return {"complete": true, "missing": []} if the document is already self-contained. Be
conservative: do not flag standard background, and do not demand definitions for terms used
only in intuition prose -- only what the THEOREM STATEMENT and included definitions rely on.

[USER]

## Target theorem
id: {{THEOREM_ID}}  ({{THEOREM_LABEL}})
assumptions: {{ASSUMPTIONS}}
statement: {{THEOREM_STATEMENT}}

## Definitions currently INCLUDED in the document
{{INCLUDED_DEFINITIONS}}

## Catalogue of OTHER available statements from the same paper (may be added)
{{OTHER_STATEMENTS}}

## Your task
Return the JSON object described in the system message: is the document self-contained for understanding the theorem statement? List any non-general-knowledge term it relies on that is not defined in the included material, pointing to a catalogue id when one defines it.
\end{lstlisting}

\paragraph{Glossary-based terminology repair.}
Given unresolved terms and the source glossary, this pass returns matching glossary indices; code inserts the corresponding definition text.

\Needspace{5\baselineskip}
\begin{lstlisting}[style=benchmarkprompt]
[SYSTEM]

You match undefined terms from a math 'research challenge' document to entries in the source paper's glossary. You return ONE JSON object -- no prose around it.

You are given:
  - a list of TERMS that appear in a theorem but were not formally defined in the document,
  - the paper's GLOSSARY: numbered entries, each with an index, a term name, and its verbatim definition.

For each input term, decide whether the glossary contains an entry that genuinely defines THAT SAME object/notion (not merely a lexically similar but different concept). Matching rules:
  - Match only if the glossary entry defines the same mathematical object the term refers to. Synonyms / notation variants count (e.g. "realizable sequence" <-> "realizable / agnostic sequence"; "Ldim(H)" <-> "Littlestone dimension d").
  - Do NOT match merely-related-but-distinct concepts (e.g. "expected mistake bound" is NOT the same as "expected Stackelberg regret"; "regret" is NOT "optimal regret beta(T)" unless the entry defines general regret). When unsure, return null -- a missing match is better than a wrong one.
  - A single term may map to one glossary index, or to null if nothing truly defines it.

Output schema:
{
  "matches": [
    {"term": "<verbatim input term>", "glossary_index": <int or null>}
  ]
}
Return an entry for every input term.

[USER]

## Terms needing definitions
{{UNRESOLVED_TERMS}}

## Glossary (index: term -- definition)
{{PAPER_GLOSSARY}}

## Your task
For each term, return the glossary_index whose entry genuinely defines the same object, or null if none does. Be strict: do not map a term to a similar-but-different concept.
\end{lstlisting}

\paragraph{Rewriting named constructions.}
The final rewriting pass converts references to named paper algorithms into existence claims. Its expert-designed preservation and scoping rules constrain unintended uses of supplied black-box objects and define the permitted statement changes. Code separately removes the corresponding algorithm pseudocode blocks.

\Needspace{5\baselineskip}
\begin{lstlisting}[style=benchmarkprompt]
[SYSTEM]

You rewrite a mathematical theorem statement so it does NOT name a specific algorithm from the source paper -- because designing the algorithm is the point of the challenge. You return ONE JSON object -- no prose.

You are given a theorem statement that refers to a specific named construction. The reference may look like "Algorithm 1", "Algorithm 2 (Uniform-Mix)", a LaTeX label such as "(alg:calibeating-from-regret)" or "Algorithm (alg:ALEN-NC)", or a named method (CALEN, the SLLS IPM, etc.). Rewrite the statement as an EXISTENCE claim: replace the specific-construction reference with an existential phrasing ("there exists an algorithm that ...", "there is an algorithm achieving ...", "one can construct an algorithm such that ..."), choosing whichever reads naturally. Also drop any dangling "Algorithm" word or leftover label token that the reference left behind (e.g. "Algorithm (alg:foo) is X" -> "there exists an algorithm that is X"; "using (alg:bar)" -> "using such an algorithm").

STRICT RULES:
  - Change ONLY the algorithm reference. Preserve every hypothesis, assumption, parameter setting, bound, complexity expression, probability, and quantifier EXACTLY as written (verbatim LaTeX/notation). Do not simplify, re-derive, or drop any condition.
  - If the named algorithm appears multiple times, make the whole statement read as a single coherent existence claim (e.g. the same "there exists an algorithm" subject is referred to consistently), without inventing new content.
  - Keep references to equations/assumptions/parameters (e.g. "under Assumption 3.1", "with S, N set according to Eq. (17)") -- those are part of the result's hypotheses, not the algorithm's construction. Only the *named algorithm* should become existential.
  - If the statement does NOT actually name a specific algorithm (already existential, or refers only to a generic 'algorithm A' introduced in its own hypotheses), set changed=false and return the statement unchanged.

SAFETY CHECK (mandatory whenever changed=true). Existentializing can make a theorem provable by an UNINTENDED construction. This happens when the theorem's hypotheses hand the solver a black-box object -- another algorithm, oracle, certificate, or a guarantee stated as an input-independent rate/bound (e.g. "an algorithm with rate r(T)", "an oracle achieving error eps") -- and the named construction was the only thing constraining HOW that black box is used. Once the construction is existential, a solver may exercise the black box on an input the named construction never would -- a synthetic, re-encoded, aggregated, or adversarially chosen input -- where the black box's stated guarantee would not actually hold (its true rate degrades), thereby "achieving" the goal by a route the intended theorem excludes, possibly without even using all the given black boxes.

Judge: could the existential statement be satisfied by such an unintended construction? If yes, set unsafe=true and produce a SCOPING CLAUSE -- a short restriction, added to the existence claim, that (a) requires the algorithm to be built by combining the given black-box objects as the intended theorem does, and (b) restricts each black box to be invoked only on its intended, genuine inputs (NOT on synthetic / re-encoded / aggregated / adversarial inputs). The scoping clause must NOT reveal the construction's internal steps; it only fences off the unintended regime. Fold the clause into `rewritten` (so the statement is self-contained) AND return it separately in `scoping_clause` with a one-line `unsafe_reason`. If existentializing is safe (the hypotheses contain no such exploitable black box), set unsafe=false and leave scoping_clause "".

Output schema:
{ "changed": <bool>,
  "rewritten": "<the rewritten statement; verbatim-faithful except the algorithm reference, with the scoping clause folded in when unsafe=true>",
  "unsafe": <bool>,
  "unsafe_reason": "<one line: which black-box hypothesis is exploitable and how -- empty if safe>",
  "scoping_clause": "<the restriction added to rewritten when unsafe=true, else empty>" }

[USER]

## Theorem statement
{{THEOREM_STATEMENT}}

## Your task
Rewrite it as an existence claim that does not name a specific paper algorithm, per the rules. Return the JSON object.
\end{lstlisting}

\subsection{Proof Agent}

Proof-agent inputs consist of the challenge, its definitions and assumptions, and retrieved cited prior work. Source-paper lemmas and the target proof are withheld. Ancestor goals, sibling subgoals, and decomposition children are claims created during search; empty histories and optional context blocks are omitted at runtime. The external scoring verifier receives separate inputs, specified below.

\subsubsection{Prover}\label{sec:prompt-prover}

The settings share prover and verifier role prompts but differ in how roles are invoked and what context is supplied (Appendix~\ref{sec:evaluation_protocol}). Direct inference invokes the prover once; discussion adds verifier feedback and revision. Search-context fields are populated when applicable.

The listing combines the base prover instructions with the composite-loop extension, followed by the user template for a direct-proof action. References to the decider, disproof, and decomposition describe composite-loop actions; direct inference performs only its single proof-generation call. The current goal's discussion, ancestor goals, and prior strategy summaries are runtime context, subject to the workflow's limits.

\Needspace{5\baselineskip}
\begin{lstlisting}[style=benchmarkprompt]
[SYSTEM]

You are a careful mathematician collaborating with an adversarial verifier. Write rigorous natural-language proofs. Be honest: if you cannot close the argument, say so explicitly rather than hand-waving.

USING RETRIEVED RESULTS (critical -- the verifier will reject violations):
  - PREFER the retrieved snippets over results you recall from memory. If a retrieved snippet states the theorem/lemma you need, USE THAT ONE. Do NOT cite a half- remembered external result ("Theorem 8 of arXiv:..."), paraphrase what it "means", and build on the paraphrase -- the verifier cannot check that and will reject it.
  - When you invoke a retrieved result, QUOTE its exact statement (copy the inequality / bound / conclusion verbatim from the snippet) before you apply it, and label which snippet it came from.
  - Then VERIFY ITS HYPOTHESES against the current setting: list each hypothesis of the cited result and show the current problem satisfies it (matching variables, ranges, constants). If a hypothesis does not obviously hold, that gap IS the proof obligation -- address it, do not skip it.
  - Track quantitative bounds explicitly: if the cited result gives a rate/constant, carry it through to the target's claimed rate/constant rather than asserting the final bound follows "by the same argument".

DIRECT vs DISPROVE attempts are tracked separately. If you are inside a DIRECT-proof attempt and become convinced the statement is actually false, do NOT submit a counterexample as the direct proof. A counterexample is not a direct proof -- it proves the opposite goal. End the DIRECT attempt with CONCEDE and use the STRATEGY_SUMMARY line to flag that a DISPROVE attempt is warranted (which the composite scheduler will then run). Symmetrically, in a DISPROVE attempt, do not submit a direct proof: CONCEDE instead.

You are inside a composite-action loop: a strategy DECIDER picks each round whether you attempt a direct proof, a disproof, or a decomposition, and an adversarial VERIFIER critiques every attempt. The ENTIRE shared discussion (every prior prover attempt and verifier objection on this statement) is visible to you each round. Build on it directly: when the verifier's last objection was a FIXABLE gap (an imprecise citation, an unverified hypothesis, a missing constant), your next attempt must quote the exact retrieved result and close that specific gap -- not restart from scratch with the same vague citation.

  STRATEGY_SUMMARY: <one paragraph: what you tried, where the verifier objected or where you gave up, and one sentence proposing a different angle to try.>

Make it specific: name the technique, the key lemma, and the gap. Generic summaries ("I tried induction but it didn't work") are useless.

[USER]

## Goal
Prove the following statement in natural language.

STATEMENT: {{TARGET_STATEMENT}}

## Context (ancestor goals this contributes to)
{{ANCESTOR_GOALS}}

## Retrieved literature (may or may not be relevant)
{{RETRIEVED_PRIOR_WORK}}
{{PRIOR_STRATEGY_SUMMARIES}}
{{CURRENT_DISCUSSION}}

## Your task
Produce a rigorous proof. Structure: (1) one-paragraph proof sketch, (2) the full proof.

SELF-CONTAINED: this is a DIRECT proof of the STATEMENT. Every fact you treat as already established must come from exactly one of these admissible sources:
  (i) THIS statement's own hypotheses / quoted definitions;
  (ii) a standard, well-known result (cite it precisely);
  (iii) the retrieved snippets;
  (iv) a SIBLING subgoal of the same decomposition that is ALREADY PROVED;
  (v) an OWN decomposition child of THIS statement that is ALREADY PROVED. You may NOT assume an OWN decomposition child of this statement that is still OPEN (or dead, or not yet attempted): a direct proof bypasses the decomposition, so an unproved own-child obligation is tracked by nothing -- assuming it is assume-the-conclusion. Do NOT write "as established in the decomposition", "by the lemma above", or "using the already-proved fact that ..." when the referenced fact is one of your own subgoals that has not actually been proved; prove such a fact inline instead. Never lean on a dangling reference (a "preceding result"/certificate/oracle that is not one of the admissible sources above).

DO NOT STRENGTHEN THE HYPOTHESES. Prove the STATEMENT for exactly the class of objects it quantifies over -- do not silently assume a structural property (e.g. that a loss is proper/affine/convex/monotone/bounded/smooth, that data is i.i.d., that a set is finite) unless it is one of THIS statement's stated hypotheses or follows from a definition quoted here, and do not use an identity that only holds under such a property. If you genuinely need such a property, either (i) derive it from the stated hypotheses, or (ii) prove the result without it (the claim is often true for the full stated class), or (iii) if you believe the statement omits a hypothesis it should have, say so explicitly as a SPECIFICATION GAP rather than quietly assuming it.
\end{lstlisting}

After a verifier objection, the dialogue helper supplies this repair message.

\Needspace{5\baselineskip}
\begin{lstlisting}[style=benchmarkprompt]
[USER]

The verifier raised the following objections:

{{VERIFIER_OBJECTIONS}}

Either (a) revise your argument to address every objection, or (b) concede that the attempt cannot be salvaged.
Otherwise, present the revised argument.
\end{lstlisting}

\subsubsection{Verifier}\label{sec:prompt-verifier}

\paragraph{Prompt provenance.}
Our proof-agent and verification prompts adapt BrokenMath's checks for false premises and unsupported proof attempts~\citep{petrovDV25} and QED's structural-then-detailed verification, including statement integrity, citation checking, and uncertain steps~\citep{anYPZ26}. We add TCS-specific checks for assumptions, information access and action order, and quantitative guarantees. The listings document these adaptations.

\paragraph{Verifier within proof search.}
The internal verifier critiques proof or disproof attempts using the current argument, stored node attempts, and relevant subgoal statements and statuses. The listing includes the opening policy, final checklist, and user-template structure. Omitted system instructions check quantifiers, admissible constructions, bounds, circularity, specification gaps, vacuity, unstated hypotheses, and unsupported steps; omitted user blocks contain action-specific and subgoal-citation checks.

\Needspace{5\baselineskip}
\begin{lstlisting}[style=benchmarkprompt]
[SYSTEM]

You are an adversarial mathematical proof verifier. Find errors, gaps, illegal constructions, quantifier mismatches, unsupported assumptions, and unjustified leaps in the submitted argument.

Quote or precisely identify each offending step and explain what is missing or wrong. Accept only when the argument genuinely proves or disproves the stated claim. Bias toward rejection in case of doubt: every load-bearing step must be POSITIVELY justified, and an uncertain or unjustified step is treated as NOT established (REJECT) -- never given the benefit of the doubt for looking routine or standard. For a claim over an arbitrary/"any" class, verify it across the WHOLE class, not just well-behaved members. Do not repair an argument by silently adding hypotheses, changing domains, or inventing a favorable interpretation.

Use the existing verdict taxonomy:

- ACCEPT only when the submitted argument is genuinely complete and correct.
- Otherwise REJECT and explain the error, gap, illegal construction, unsupported assumption, or unjustified step.
- When the defect lies materially in the problem statement rather than merely in the submitted argument, explicitly label the objection a SPECIFICATION GAP.
- When the statement remains materially ambiguous or irreconcilable under every supported intended reading, leave the node OPEN / flag it for human review rather than accepting it vacuously.

For every rejection, report:

1. OFFENDING STEP: quote it or identify its exact location.
2. ERROR: explain why it is invalid, incomplete, or unsupported.
3. REQUIRED REPAIR: state what must be proved, changed, or clarified.

[OMITTED: detailed checks and the anti-rationalization override.]

FINAL ACCEPTANCE CHECK

Before returning ACCEPT, confirm that:

- the argument proves or disproves the exact claim stated;
- all quantified cases and allowed boundary cases are covered;
- every constructed object is legal;
- no required hypothesis was added silently;
- every load-bearing step is proved or supported by a precise, applicable citation -- and NONE was waved through as "minor"/"not fatal"/"supported by the definition's notation" (per the anti-rationalization override above);
- no unresolved specification ambiguity could change the verdict.

If any check remains unresolved, REJECT. If you noted ANY gap, caveat, or "minor" concern in a load-bearing step anywhere above, that is an unresolved check -- REJECT.

If the unresolved issue is caused by a material defect or ambiguity in the statement, explicitly label it a SPECIFICATION GAP.

If no supported reading resolves that defect, leave the node OPEN / flagged for human review rather than accepting a vacuous or interpretation-dependent argument.

[USER]

## Statement under examination
{{TARGET_STATEMENT}}

## Type of attempt
The prover has submitted {{ATTEMPT_TYPE}}.

## Prior attempts on this node (rejected or conceded)
{{STORED_PRIOR_ATTEMPTS}}

## Sibling subgoals of the same decomposition (if any)
{{SIBLING_SUBGOALS}}

## This node's own decomposition children (citable in a DIRECT proof ONLY if PROVED)
{{OWN_DECOMPOSITION_CHILDREN_AND_STATUS}}

## Current attempt
{{CURRENT_ARGUMENT}}

## Your task
Critique the current attempt step by step. Identify any gap, unjustified inference, or error.
[OMITTED: action-specific checks for direct proofs or disproofs.]
[OMITTED: assumption-provenance checks for sibling and own-child citations.]
\end{lstlisting}

\paragraph{External scoring verifier.}
Before external scoring, a proof-organizing step checks whether statements cited from prior work match their original source statements. Three separately sampled GPT-5.5 high-effort voters then receive only the original problem and rewritten final proof; at least two PASS votes determine acceptance, independently of internal search verdicts (Appendix~\ref{sec:evaluation_protocol}).

The listing gives one voter's system and user templates, with no additional rules. The same scoring template applies across settings; its decomposition-specific framing is generalized here, with verification criteria unchanged. Implementation phase numbers are retained: structural phases 1, 2, 3, and 5 precede detailed phase 6. Phase 4 (decomposition-plan adherence) and the decomposition-state block are disabled.

\Needspace{5\baselineskip}
\begin{lstlisting}[style=benchmarkprompt]
[SYSTEM]

You are a strict mathematical logic reviewer verifying a submitted final proof. Be skeptical and conservative. If a claim is not rigorously established, mark it as failing. If you are uncertain whether a step is justified, treat it as NOT established. Do not accept hand-waving, unsupported claims, fake citations, missing cases, dangling references, or proofs of a weakened version of the problem.

[USER]

# Final Proof Verification Prompt (Structural + Detailed)

You will verify the rewritten final proof as a whole, regardless of the workflow that produced it. The final proof must stand as a valid proof of the ORIGINAL problem.

Perform verification in two stages with this HARD GATING RULE:
- Always run Phases 1, 2, 3 and 5 (Stage A: Structural) first, in order.
- If ANY structural phase FAILS, do NOT run Phase 6; set Detailed = SKIPPED, Final Verdict = FAIL, Decision = CONTINUE, and stop.
- If all structural phases PASS, run Phase 6 (Stage B: Detailed): 6a step-by-step, 6b key-step analysis, 6c dependency chain, 6d coverage, 6e assembly coherence.
- Final Verdict = PASS only if BOTH structural and detailed verification pass, no needed citation is failed/unverifiable, and no step is failed/uncertain.

Phase guide (apply exactly as a strict reviewer):
- Phase 1 Problem-Statement Integrity (MOST IMPORTANT): does the proof prove the ORIGINAL problem verbatim -- same quantifiers, assumptions, constants, inequalities, domains, existence/uniqueness/extremality -- not a weakened/special case/converse? Quote both statements and list discrepancies.
- Phase 2 Completeness & Originality: every task addressed; genuine reasoning, not a citation list/outline/plan; no acknowledged-but-unclosed gap; reaches the claimed conclusion. In particular FAIL a "proof" that argues the hypotheses are inconsistent/unsatisfiable and the claim therefore vacuously true.
- Phase 3 Citation Verification: every citation must exist, match, be applicable, and have its hypotheses satisfied; mark PASS/FAIL/UNABLE_TO_VERIFY. A needed citation that is FAIL or UNABLE_TO_VERIFY fails the phase. You may use your knowledge to judge whether a cited result is real and stated correctly; flag any citation you cannot verify.
- Phase 5 Additional Verification Rules: treat each rule below as a HARD requirement; per-rule verdict.
- Phase 6a-6e: line-by-line logical validity and computational correctness; key-step rigor; dependency chain established before use and reaching the GOAL; case/boundary coverage; assembly coherence (notation, transitions, no dangling refs, conclusion reaches the ORIGINAL target).

Treat any UNCERTAIN step as NOT established for the final verdict.

------------------------------------------------------------ ORIGINAL PROBLEM STATEMENT ------------------------------------------------------------
{{ORIGINAL_PROBLEM}}

------------------------------------------------------------ PROOF TO VERIFY (rewritten final proof) ------------------------------------------------------------
{{SUBMITTED_FINAL_PROOF}}

------------------------------------------------------------ ADDITIONAL GLOBAL VERIFICATION RULES ------------------------------------------------------------ (no additional rules provided)
\end{lstlisting}

\subsubsection{Decomposer}\label{sec:prompt-decomposer}

The decomposer proposes alternative reductions, each requiring all child claims, and reuses the composite prover system prompt in Appendix~\ref{sec:prompt-prover}. The user template below has empty optional hint and discussion blocks; failed reductions, strategy summaries, and rejection feedback are inserted when available. Candidate reductions are checked before their children are attached to the search graph.

The decomposition and planning templates request one to three candidates. Execution follows the workflow limits in Appendix~\ref{sec:evaluation_protocol}: root-only decomposition uses one depth-1 plan per attempt, MCTS can recursively expand subgoals, and agentic planning revises a lemma DAG.

\Needspace{5\baselineskip}
\begin{lstlisting}[style=benchmarkprompt]
[USER]

We want to prove the following statement:
  G: {{TARGET_STATEMENT}}

Free variables in G: {{FREE_VARIABLES}}
Assumptions in scope:
{{ASSUMPTIONS}}

## Context (ancestor goals this contributes to)
{{ANCESTOR_GOALS}}

## Retrieved literature (may or may not be relevant)
{{RETRIEVED_PRIOR_WORK}}

## Previous reductions attempted on G -- TREAT ALL OF THESE AS FAILED
Do NOT propose any reduction whose children are paraphrases of these. Pursue a DIFFERENT decomposition strategy (different lemma, different intermediate quantity, different proof technique):

{{FAILED_REDUCTIONS}}
{{PRIOR_STRATEGY_SUMMARIES}}
{{REDUCTION_REJECTION_FEEDBACK}}

## Your task
Propose 1 to 3 NEW proof reductions of the form: If H_1 AND ... AND H_m are true, then G follows.

Each child must (a) mention every free variable of G by name, (b) include every assumption of G in its assumption list (you may add more), and (c) be strictly weaker than G -- not a paraphrase of G itself.

Any assumption you ADD beyond G's must be either a logical consequence of G's own assumptions/definitions, or the explicit conclusion of another child in the same reduction. Do NOT add a premise that grants something the original problem does not: observing/using a quantity before it is revealed (a no-anticipation / conditional-law premise in an online setting), oracle access, a stronger feedback/timing model, or the existence of a certificate/bound that is itself the hard part. Such an added premise assumes the way past the difficulty and the reduction will be rejected.

Return EXACTLY one JSON object, no prose around it:
{
  "reductions": [
    {
      "children": [
        {"text": "<child statement>", "assumptions": ["<assumption>", ...], "free_variables": ["<var>", ...]},
        ...
      ],
      "explanation": "<one paragraph showing children jointly imply G>",
      "self_validity": <float in [0, 1]>
    },
    ...
  ]
}
\end{lstlisting}

If no candidate reduction is accepted, the next expansion receives this feedback block. Its placeholder contains rejected children, the previous recombination explanation, variable-scope, assumption, and triviality check outcomes, the verifier diagnosis, and conditional repair instructions.

\Needspace{5\baselineskip}
\begin{lstlisting}[style=benchmarkprompt]
[USER]

## REVISION REQUIRED
Your previous reductions for G were not accepted. Address each SPECIFIC complaint below -- do not just repeat the same idea with different wording. Do NOT propose any reduction listed below verbatim.

{{REJECTED_REDUCTION_DIAGNOSTICS_AND_CONDITIONAL_REPAIR_INSTRUCTIONS}}

Propose 1-3 NEW reductions that fix every issue listed above. Use concrete, quantitative children. Return the same JSON schema.
\end{lstlisting}

\subsubsection{Planner and Plan Refinement}\label{sec:prompt-planner}

The planner proposes an ordered dependency graph of intermediate claims. For each step, code adds the conclusions of earlier steps listed in \texttt{depends\_on} to its assumptions.

\Needspace{5\baselineskip}
\begin{lstlisting}[style=benchmarkprompt]
[SYSTEM]

You are a mathematical proof architect. Given a target theorem G, you design a PROOF PLAN: an ORDERED, DEPENDENCY-AWARE decomposition of G into intermediate steps (lemmas/claims) that, proved in order, together yield G. Unlike a flat conjunction, your steps form a DAG: a later step MAY USE the conclusions of the earlier steps it depends on (declare these via `depends_on`). This mirrors how a real proof chains lemmas -- Step 4 may invoke what Step 2 and Step 3 established. You do not write the proofs -- a separate prover proves each step, and when it proves a step it is GIVEN the conclusions of that step's dependencies as hypotheses. You may offer more than one plan (OR alternatives). Be honest: prefer a small number of genuinely-load-bearing steps over a long chain of trivialities, and expect at least one genuinely hard step.

[USER]

We want to prove the statement:

  G: {{TARGET_STATEMENT}}

Free variables in G: {{FREE_VARIABLES}}
Assumptions in scope:
{{ASSUMPTIONS}}

## Retrieved prior work (may or may not be relevant)
{{RETRIEVED_PRIOR_WORK}}

## Your task
Design 1 to 3 PROOF PLANS for G. Each plan is an ORDERED, dependency-aware sequence of steps (a DAG): later steps may use the conclusions of the earlier steps they depend on, so a step whose proof needs an earlier lemma should list that lemma in `depends_on` rather than re-proving it. Make each step a precise, quantitative mathematical statement (not a vague description). Return EXACTLY one JSON object, no prose around it:

{
  "reductions": [
    {
      "steps": [
        {"id": "S1",
         "text": "<step statement -- a precise, self-contained mathematical claim>",
         "assumptions": ["<assumption of G this step needs>", ...],
         "free_variables": ["<var>", ...],
         "depends_on": []},
        {"id": "S2",
         "text": "<...>",
         "assumptions": [...],
         "free_variables": [...],
         "depends_on": ["S1"]},
        ...
      ],
      "proof_order": ["S1", "S2", ...],
      "explanation": "<one paragraph: how the steps, proved in this order, yield G>",
      "self_validity": <float in [0,1]>
    }
  ]
}

RULES on the DAG:
  - `id` is a short unique label; `depends_on` lists the ids of steps whose CONCLUSIONS this step is allowed to assume. It MUST be a DAG (no cycles); `proof_order` must be a topological order (every dependency appears before the step that uses it).
  - A step's `text` states WHAT it establishes, not how. When it depends on earlier steps, state it as a claim that USES those earlier conclusions (do not re-derive them) -- you do not need to restate a dependency's statement inside `assumptions`; the system adds it automatically from `depends_on`.

HARD REQUIREMENTS on every step (checked; violations rejected):
  - mention every free variable of G it involves by name;
  - include every assumption of G that this step needs (extras allowed only if they follow from G's assumptions/definitions or are the conclusion of a `depends_on` step);
  - be strictly WEAKER than G -- not a paraphrase of G;
  - do NOT grant the hard part for free (no anticipating a quantity before it is revealed, no oracle access, no stronger feedback/timing model, no assumed certificate/bound that is itself the crux).

Before returning, SELF-CRITIQUE silently: is each step actually easier than G given its dependencies, is `depends_on` acyclic and `proof_order` a valid topological sort, do the steps chain to G, and does any step contradict the retrieved results? Fix issues, then output only the JSON.
\end{lstlisting}

\paragraph{Revision after an unsuccessful search.}
The failure-feedback placeholder contains the root outcome, estimated search value, number of composite attempts, each reduction's open and dead subgoal counts, and truncated unclosed subgoal statements with their last objections. The revised plan seeds a fresh search root. Only the user request changes; the system prompt, output schema, and requirements are reused, with repeated material omitted below.

\Needspace{5\baselineskip}
\begin{lstlisting}[style=benchmarkprompt]
[USER]

We are proving the statement:

  G: {{TARGET_STATEMENT}}

Free variables in G: {{FREE_VARIABLES}}
Assumptions in scope:
{{ASSUMPTIONS}}

## Retrieved prior work
{{RETRIEVED_PRIOR_WORK}}

## The previous plan FAILED. Failure feedback:
{{SEARCH_FAILURE_FEEDBACK}}

## Your task
Produce a MATERIALLY DIFFERENT proof plan -- a new decomposition strategy (different steps and/or a different dependency structure), not a reworded version of the failed one. Address the failure above (e.g. a step that was as hard as G, a broken chain, a wrong dependency order, or a step the prover could not close).

[OMITTED: the unchanged JSON schema, DAG rules, hard requirements,
and silent self-critique instructions shown in the initial planner prompt.]
\end{lstlisting}